\documentclass{article}

\PassOptionsToPackage{numbers, compress}{natbib}
\usepackage[preprint]{neurips_2026}

\usepackage[utf8]{inputenc} 
\usepackage[T1]{fontenc}    
\usepackage{hyperref}       
\usepackage{url}            
\usepackage{booktabs}       
\usepackage{amsfonts}       
\usepackage{nicefrac}       
\usepackage{microtype}      
\usepackage{xcolor}         

\usepackage{graphicx}
\usepackage{amsmath}    
\usepackage{subcaption}

\usepackage{booktabs}
\usepackage[table]{xcolor}
\usepackage{graphicx}
\usepackage{tabularx}
\usepackage{array}
\usepackage{threeparttable}
\usepackage{multirow}
\usepackage{subcaption}
\usepackage{adjustbox}
\usepackage{makecell}

\title{Reroute, Don't Remove: Recoverable Visual Token Routing for Vision-Language Models}

\author{%
  Cheng-Yu Yang \\
  National Yang Ming Chiao Tung University\\
  \texttt{elm855220@gmail.com} \\
  \And
  Shao-Yuan Lo \\
  National Taiwan University \\
  \texttt{sylo@csie.ntu.edu.tw} \\
  \AND
  Yu-Lun Liu \\
  National Yang Ming Chiao Tung University \\
  \texttt{yulunliu@cs.nycu.edu.tw} \\
}

\begin{document}

\maketitle


\begin{abstract}
    Vision-language models (VLMs) project images into hundreds to thousands of visual tokens, making decoder inference expensive in both attention computation and KV-cache memory. Existing visual-token reduction methods largely follow a \emph{rank-and-remove} paradigm: they score visual tokens, keep a compact subset, and permanently discard the rest. We show that this irreversible action is fragile because visual-token importance changes across decoder depth; tokens ranked low at one stage may become relevant in later layers, especially for grounding-sensitive queries. We propose \textbf{Reroute}, a training-free plug-in that replaces removal with recoverable routing. At each routing stage, selected vision tokens pass through decoder blocks, while deferred tokens bypass the stage and re-enter the candidate pool at the next routing decision. Reroute reuses existing attention-score ranking rules and stage-wise schedules, preserving the theoretical TFLOPs and KV-cache budget class of the pruning method it augments. Across FastV, PDrop, and Nüwa variants on LLaVA-1.5 and Qwen backbones, reroute improves grounding under aggressive token reduction while maintaining general VQA performance. These results suggest that VLM token reduction should not be viewed only as irreversible pruning, but also as recoverable routing.
    The code can be found here: \url{https://github.com/elmma/mllm-reroute/}.
\end{abstract}

\begin{figure}[t]
    \centering
    \vspace{-3mm}
    \includegraphics[width=1.0\linewidth]{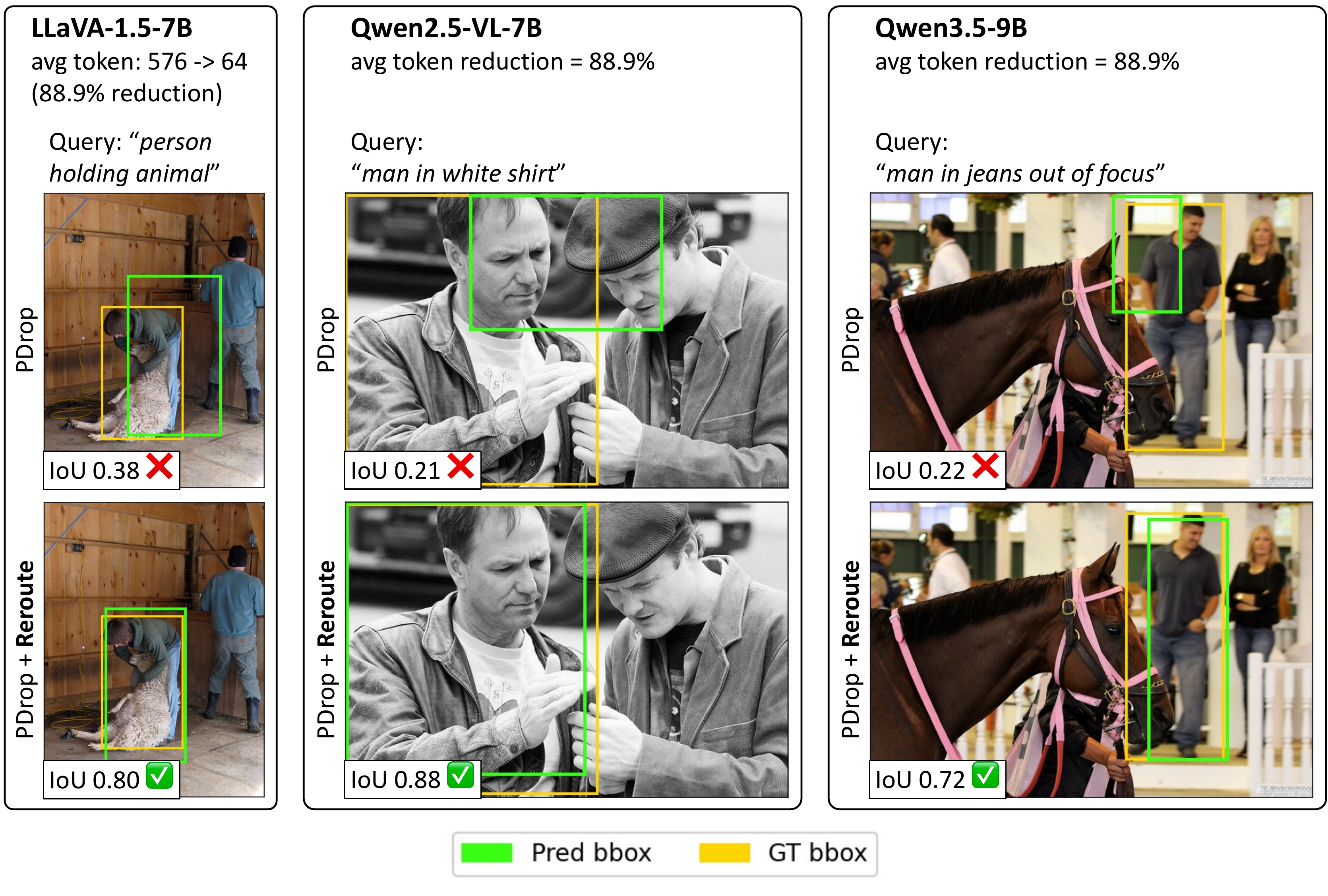}
    \vspace{-6mm}
    \caption{\textbf{Reroute, don't remove.}
    Under aggressive 88.9\% visual token reduction (576→64 visual tokens), conventional pruning permanently discards visual evidence and grounding collapses (top rows, IoU < 0.4). Replacing irreversible removal with our recoverable routing, using the same scorer (PDrop) and the same token budget, we can recover grounding accuracy across three backbones spanning early-generation MLLM (LLaVA-1.5-7B), native dynamic resolution (Qwen2.5-VL-7B), and hybrid Mamba-Transformer (Qwen3.5-9B). Reroute is a training-free plug-in that improves grounding at matched theoretical cost.
    }
    \label{fig:teaser}
    \vspace{-2mm}
\end{figure}

\vspace{-2mm}
\section{Introduction}
\vspace{-2mm}
Vision-language models (VLMs)~\cite{liu2023visual,bai2023qwen} project images into hundreds to thousands of tokens that are fed to a language-model decoder, making inference cost grow quadratically with attention and linearly with KV-cache. Reducing the number of visual tokens that participate in decoder computation has therefore become a key lever for deploying VLMs at scale~\cite{chen2024image,xing2024pyramiddrop,huang2026n}.

Existing methods follow a common \emph{rank-and-remove} recipe. Vision-encoder-side approaches~\cite{yang2025visionzip,shang2025llava} compress tokens before they enter the decoder. Decoder-side methods intervene inside the LLM: FastV~\cite{chen2024image} makes a single-shot pruning decision at an early layer using text-to-vision attention; PDrop~\cite{xing2024pyramiddrop} generalizes this to a progressive multi-stage schedule; FEATHER~\cite{endo2025feather} and Nüwa~\cite{huang2026n} refine the scoring rule to better preserve spatial evidence. Across these methods, the underlying \emph{action} is identical. Once a visual token is deemed unimportant, it is removed from the sequence and cannot contribute to any later layer.

\begin{figure}[t]
    \centering
    \vspace{-3mm}
    \includegraphics[width=1.0\linewidth]{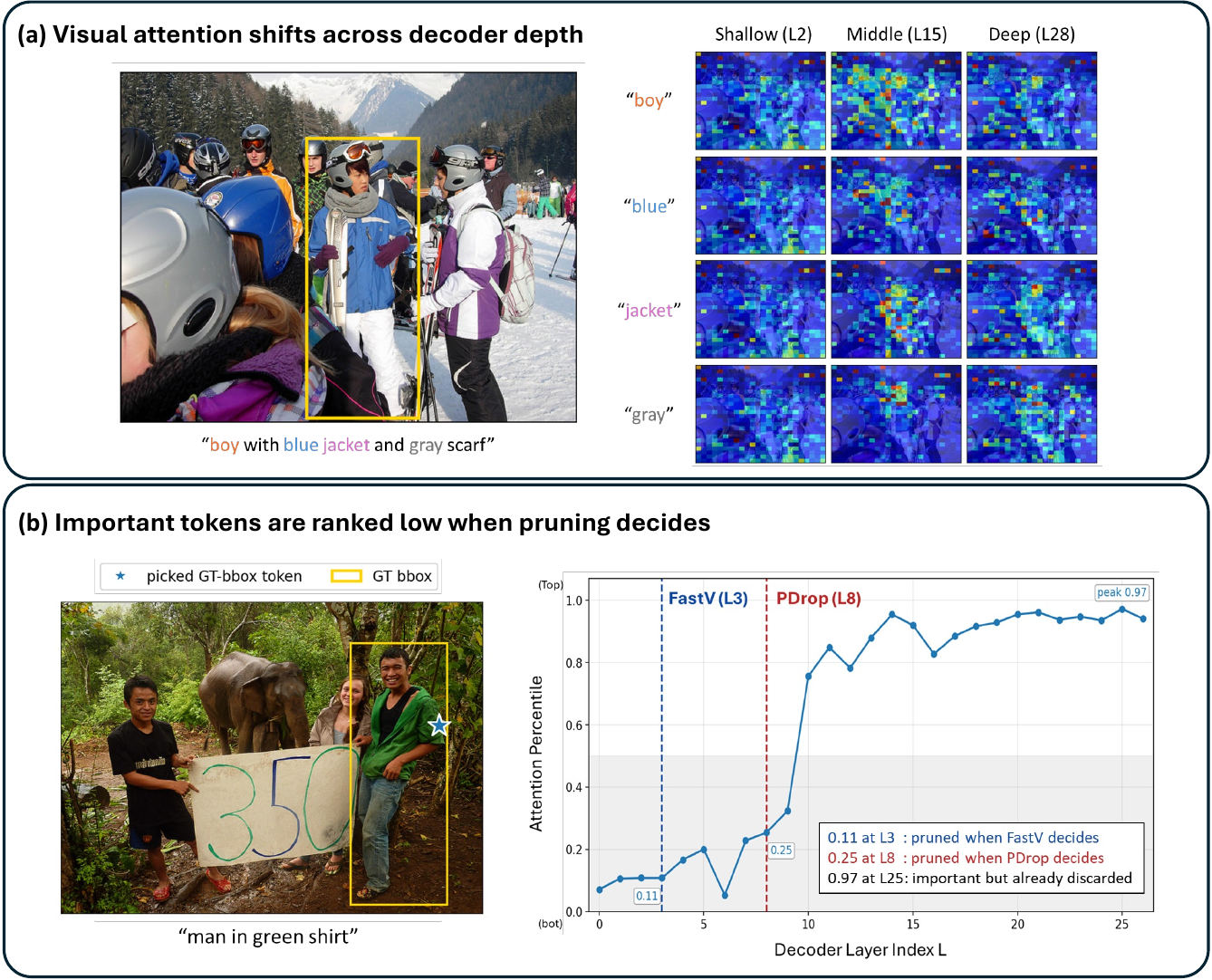}
    \vspace{-6mm}
    \caption{\textbf{Why reroute? Important visual tokens are ranked low when pruning decides.} (a) On a single RefCOCO image, attention from each query word (``boy'', ``blue'', ``jacket'', ``gray'') to visual tokens drifts substantially across decoder depth. Shallow layers attend diffusely, while target-relevant regions only emerge at middle and deep layers. (b) Tracking one ground-truth token ($\star$) inside the GT bbox of ``man in green shirt'': its attention percentile is 0.11 at L3 (where FastV drops) and 0.25 at L8 (where PDrop drops), so both methods irreversibly remove it. Yet its importance climbs to 0.97 by L25, a token that is essential at deep layers but indistinguishable from background at the layers where pruning commits. Reroute keeps such tokens recoverable.}
    \label{fig:motivation}
    \vspace{-2mm}
\end{figure}

This paradigm assumes token importance is stable across decoder depth. Our analysis often shows this assumption is not true. Figure~\ref{fig:motivation}(a) tracks attention from query words to visual tokens across depth: shallow layers attend diffusely, while target-relevant regions emerge only in middle and deep layers. Figure~\ref{fig:motivation}(b) follows one ground-truth token for ``man in green shirt'': its attention percentile is \textbf{0.11 at layer 3} (where FastV decides) and \textbf{0.25 at layer 8} (where PDrop decides). Both methods irreversibly remove it, yet it climbs to \textbf{0.97 at layer 25}. As Figure~\ref{fig:teaser} shows, this irrecoverability causes outright grounding collapse under 88.9\% token reduction across three backbones, with IoU falling below 0.4.

\begin{figure}[t]
    \centering
    \vspace{-3mm}
    \includegraphics[width=1.0\linewidth]{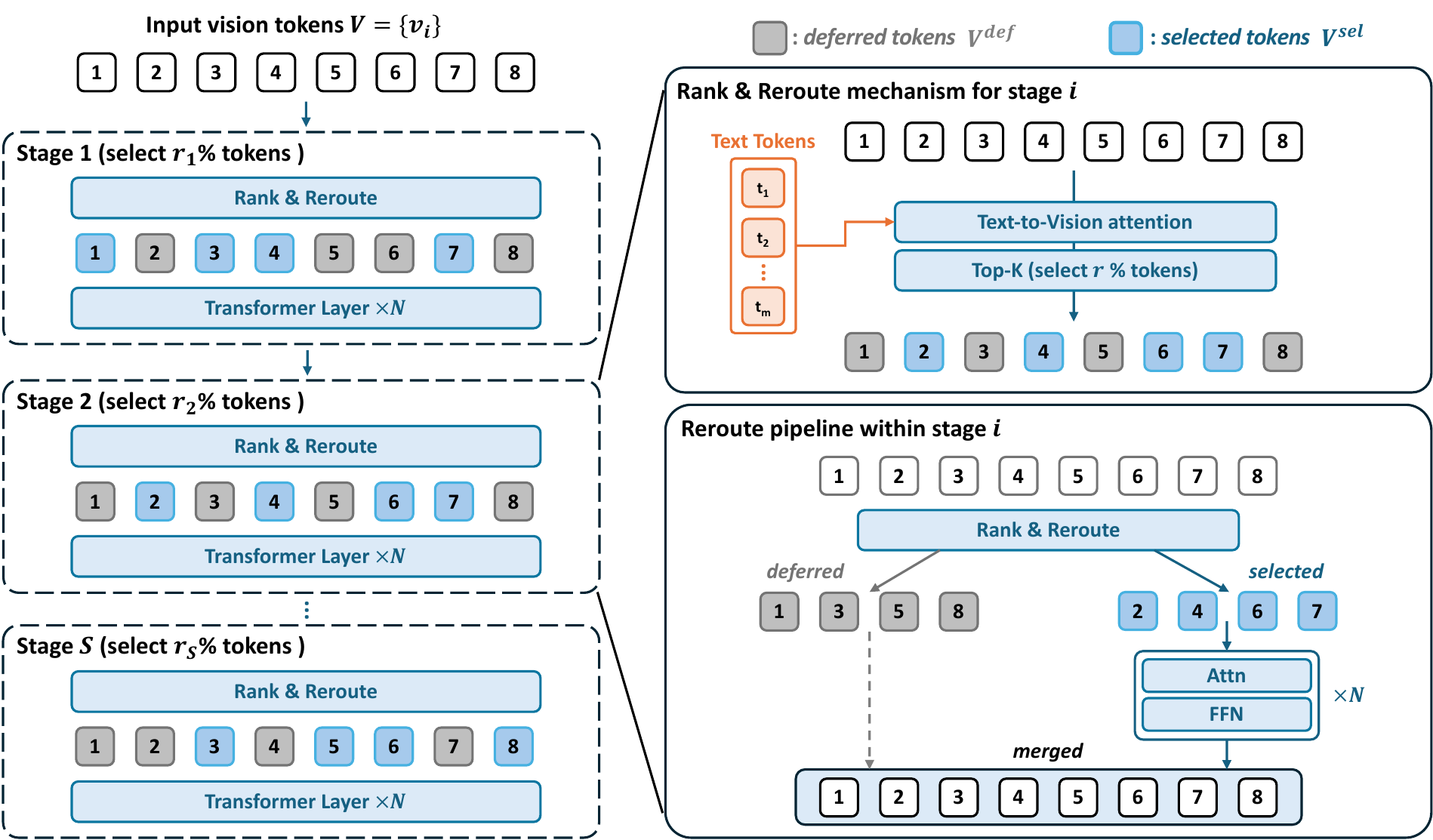}
    \vspace{-6mm}
    \caption{\textbf{Overview of Reroute.} Reroute organizes decoder-side visual token reduction as a sequence of $S$ routing stages, each consisting of a \texttt{Rank \& Reroute} operator followed by $N$ transformer layers. \emph{(Left)}: The full pipeline. Input visual tokens $V = \left\{ v_i \right\}$ pass through stages $1...S$; at each stage $i$, the operator selects a top-$r_i$\% subset to route through the transformer block, while the remaining \emph{deferred} tokens bypass it via a residual path. Crucially, deferred tokens are not removed from V. They re-enter the next stage's ranking and may be reselected. \emph{(Right)}: Within one stage, selected and deferred tokens take separate computational paths and merge back at the stage output, restoring the original sequence ordering for the next stage. The \texttt{Rank \& Reroute} operator itself reuses the existing text-to-vision attention from the current decoder layer as the ranking signal. No additional router is trained. The top-$K$ visual tokens ($K=\left\lfloor r_i \cdot \left| V \right| \right\rfloor$) are marked \emph{selected}; the rest are marked \emph{deferred}. Reroute therefore preserves the FLOPs and KV-cache budget of conventional pruning (only $\left| r_i \right|$ tokens enter Attn + FFN), while breaking the irreversibility constraint that pruning imposes on $V$.}
    \label{fig:pipeline}
    \vspace{-2mm}
\end{figure}

We argue the limitation lies not in the scorer but in the \emph{action} taken after scoring. We propose \textbf{Reroute}, a training-free plug-in that replaces irreversible removal with recoverable routing (Figure~\ref{fig:pipeline}). At each routing layer, Reroute reuses text-to-vision attention as the ranking signal and selects a top-$r$ fraction of tokens to pass through Attn+FFN. Deferred tokens are \emph{not} deleted. They bypass the stage via a residual path and re-enter the candidate pool at the next routing layer, evaluated against the \emph{full} visual-token set. A token deferred at layer 3 can be re-admitted at layer 15 if its score warrants it. Since only selected tokens enter Attn+FFN per stage, Reroute preserves the FLOPs and KV-cache budget of the pruning baseline it augments. From a conditional-computation view, Reroute is a training-free, attention-driven instantiation of mixture-of-depth~\cite{raposo2024mixture} for VLM visual tokens, in contrast to learned-router approaches like $\gamma$-MoD~\cite{luo2024gamma}. We validate Reroute as a drop-in augmentation of FastV, PDrop, and Nüwa across LLaVA-1.5-7B, Qwen2.5-VL-7B, and Qwen3.5-9B, with the largest gains in grounding-heavy and aggressive-pruning regimes.

Our contributions are as follows:
\begin{itemize}
\vspace{-1mm}
    \item \textbf{Recoverable routing formulation.} We recast decoder-side visual-token pruning as stage-wise routing with deferred-token bypass, making conventional irreversible pruning the degenerate case where re-entry is forbidden. This enables a training-free, attention-driven realization of mixture-of-depth for VLM visual tokens.
\vspace{-1mm}
    \item \textbf{Plug-in mechanism with matched theoretical efficiency.} Reroute reuses existing scorers and schedules with no extra trainable parameters, preserving the FLOPs and KV-cache budget of the pruning method it augments.
\vspace{-1mm}
    \item \textbf{Consistent gains across methods and backbones.}
Applied to FastV, PDrop, and Nüwa across three backbones, Reroute improves matched-budget pruning baselines on grounding benchmarks and preserves general VQA performance, with the largest gains under aggressive token reduction.
\vspace{-1mm}
\end{itemize}

\vspace{-2mm}
\section{Related Work}
\label{sec:related}

\vspace{-2mm}
\paragraph{Vision-encoder-side visual token reduction.}
A long line of work compresses visual information \emph{before} it enters
the decoder. ViT pruning supplies the basic primitives, score-and-drop
\cite{rao2021dynamicvit,liang2022not} and similarity-based merging
\cite{bolya2022token}. These ideas are imported into VLMs through methods
that operate on the projector or \texttt{[CLS]} head:
LLaVA-PruMerge \cite{shang2025llava}, VisionZip \cite{yang2025visionzip},
FasterVLM \cite{zhang2024cls}, and the budget-aware HiRED for
sub-image partitioning \cite{arif2025hired}. A complementary thread
redesigns the projector itself
\cite{li2025tokenpacker,cha2024honeybee,hu2024matryoshka,cai2024matryoshka,
ye2025voco,chen2024efficient,wang2025folder},
while DeepStack instead distributes high-resolution tokens across decoder
layers without enlarging the context \cite{meng2024deepstack}. Other
works prune at the patch level with training-free saliency
\cite{mahmud2024papr} or jointly compress encoder and decoder
\cite{hu2024illava}. All commit to a \emph{fixed} visual representation
before language reasoning begins. In contrast, Reroute intervenes
\emph{inside} the decoder and revisits the visual-token set at every
routing stage.

\vspace{-2mm}
\paragraph{Decoder-side visual token pruning.}
FastV initiated this line by observing that vision-attention mass collapses
after layer 2 and pruning with a single text-to-vision rule
\cite{chen2024image}. PyramidDrop generalized this to a progressive
multi-stage schedule \cite{xing2024pyramiddrop}; SparseVLM combined
pre-LLM filtering with in-LLM recycling \cite{zhang2024sparsevlm};
FEATHER traced FastV's grounding failures to spatial bias and added a
uniform-coverage safety net \cite{endo2025feather}; and N\"uwa enforces
spatial integrity, becoming the strongest grounding-oriented baseline
\cite{huang2026n}. The same \emph{score-then-remove} skeleton has been
refined along nearly every axis: divergence-based calibration
\cite{ye2025fit}, layer-adaptive budgets with KV compression
\cite{he2024zipvl}, multi-stage drops with a protected key-set
\cite{liu2024multi}, terminal visual withdrawal
\cite{lin2025boosting,wu2024accelerating}, and diversity- or similarity-aware
selection \cite{alvar2025divprune,zhang2025beyond,yang2025topv,jiang2024fopru,liu2026hiprune,chen2026evoprune}.
Trained variants predict instance- and layer-specific keep ratios
\cite{huang2024dynamic,ye2025atp,zhuang2025st3}. Across this body
of work, the post-selection action is identical: deferred tokens are
\emph{permanently removed} and cannot influence later layers. The closest
exception, Recoverable Compression, restores dropped evidence via
text-guided similarity at a single recovery point \cite{chen2025recoverable}.
We instead recast the entire decoder schedule as a sequence of
\emph{recoverable routing} decisions: a token deferred at one stage stays
in the sequence and competes for re-entry at every subsequent stage.

\vspace{-2mm}
\paragraph{Mixture-of-depth and conditional computation.}
Adaptive Computation Time \cite{graves2016adaptive} and early-exit networks
\cite{teerapittayanon2016branchynet,huang2017multi,xin2020deebert,liu2020fastbert,schuster2022confident} let inputs spend variable depth, but
exit is \emph{terminal}. Layer-skipping policies route inputs through a
subset of blocks via learned gates \cite{wang2018skipnet,wu2018blockdrop}.
Mixture-of-Depths refines this for transformers: each layer routes only
the top-$k$ tokens through its block while the rest take a residual bypass
\cite{raposo2024mixture}. $\gamma$-MoD \cite{luo2024gamma} and p-MoD
\cite{zhang2025p} extend MoD to MLLMs, and Routing Experts
\cite{wu2025routing} and Mixture-of-Recursions \cite{bae2025mixture} explore
related dynamic-routing variants; a recent bypass-and-attend study
analyzes how skipped positions interact with later attention
\cite{lawson2025learning}. Multimodal MoE methods pursue width-based
rather than depth-based sparsity \cite{lin2026moe,li2024cumo,zong2024mova}. Reroute differs in two ways. First, where these methods
\emph{train} a new router (often the model with it), our router
\emph{reuses} the text-to-vision attention scores already computed by
the decoder, making the mechanism \emph{training-free}. Second, where
they target the LLM's own tokens, we target \emph{visual} tokens, where
layer-wise importance shifts are pronounced and existing pruning practice
provides a free routing signal.

\vspace{-2mm}
\paragraph{Layer-wise attention dynamics in VLM decoders.}
The phenomenon we exploit, visual-token importance, is unstable across depth. This has been documented from several angles. FastV first noted the
post-layer-2 attention drop \cite{chen2024image}; FasterVLM and VisPruner
identified two failure modes, \emph{attention shift} and \emph{attention
dispersion} \cite{zhang2024cls,zhang2025beyond}. Massive
Activations \cite{sun2024massive} and the Visual Attention Sink
\cite{kang2025see} explain part of this through tokens that absorb
disproportionate attention without carrying matching semantics; analogous
sinks were earlier reported in pure LLMs \cite{xiao2023efficient}.
Recent VLM analyses argue for a phased shallow/middle/deep view of visual
information flow \cite{yin2025lifting}, and MCA-LLaVA traces part of the
drift to RoPE-induced positional bias \cite{zhao2025mca}. Mechanistic
studies of induction and retrieval heads provide broader context
\cite{olsson2022context,wu2024retrieval}. Rather than pruning at
a single ``best'' layer, we treat depth-dependent importance as a feature
to be \emph{exploited} and let deferred tokens re-enter when their score
warrants it.

\vspace{-2mm}
\paragraph{Efficient inference and KV-cache compression.}
Our work is complementary to KV-cache eviction
\cite{xiao2023efficient,zhang2023h2o,liu2023scissorhands,ge2023model,li2024snapkv,cai2024pyramidkv} and its multimodal extensions
\cite{wan2024look,liu2024efficient}, as well as to system-level
optimizations \cite{dao2022flashattention,kwon2023efficient}. These methods
decide which past states to keep in the cache; we decide which visual
tokens deserve full computation in the first place. Reroute can in
principle be combined with any of them.

\vspace{-2mm}
\section{Preliminaries}
\label{sec:preliminaries}

\vspace{-2mm}
\paragraph{Multimodal decoder notation.}
\label{sec:multimodal-decoder-notation}
Let $\mathcal{T}, \mathcal{V}$ index text and vision token positions. At layer $\ell$, hidden states $H^\ell = \{h_j^\ell\}_{j \in \mathcal{T} \cup \mathcal{V}}$ are updated by a standard decoder block:
\begin{equation}
H^{\ell+1} = \tilde{H}^{\ell+1} + \operatorname{FFN}_\ell(\operatorname{LN}_2(\tilde{H}^{\ell+1})), \quad \tilde{H}^{\ell+1} = H^\ell + \operatorname{Attn}_\ell(\operatorname{LN}_1(H^\ell)).
\end{equation}
Following the visual-token pruning literature, reduction targets only vision tokens; text tokens stay active throughout decoding, isolating the post-ranking action on low-ranked vision tokens.

\vspace{-2mm}
\paragraph{Stage-wise visual-token selection.}
\label{sec:stage-wise-selection}
We cast decoder-side visual-token reduction as stage-wise selection. The decoder is partitioned into $S$ stages, with stage $i$ starting at routing layer $\ell_i$ with keep ratio $r_i$. Let $\mathcal{C}_i \subseteq \mathcal{V}$ denote the candidate vision-token set at the start of stage $i$, with $\mathcal{C}_1=\mathcal{V}$. At layer $\ell_i$, a scorer (typically text-to-vision attention) ranks $\mathcal{C}_i$ and picks $\mathcal{V}_i^{\mathrm{sel}} = \operatorname{Top-K}_i(\mathcal{C}_i)$ with $K_i = \lfloor r_i \lvert \mathcal{V} \rvert \rfloor$; the deferred set is $\mathcal{V}_i^{\mathrm{def}} = \mathcal{C}_i \setminus \mathcal{V}_i^{\mathrm{sel}}$. Since text tokens remain active in every stage, the active set is $\mathcal{A}_i = \mathcal{T} \cup \mathcal{V}_i^{\mathrm{sel}}$. The reduction method is defined by the action on $\mathcal{V}_i^{\mathrm{def}}$: pruning removes them, whereas Reroute keeps them recoverable.

\vspace{-2mm}
\paragraph{Rank-and-remove as irreversible pruning.}
\label{sec:rank-and-remove-pruning}
Conventional pruning removes $\mathcal{V}_i^{\mathrm{def}}$, contracting the candidate pool to $\mathcal{C}_{i+1}^{\mathrm{prune}} = \mathcal{V}_i^{\mathrm{sel}}$. Once deferred, a vision token cannot be re-scored or re-selected in later stages. This irreversibility defines rank-and-remove pruning.

\begin{figure}[t]
    \centering
\vspace{-3mm}
    \captionsetup{skip=3pt}

    \includegraphics[
        width=0.94\linewidth,
        trim=4pt 4pt 4pt 4pt,
        clip
    ]{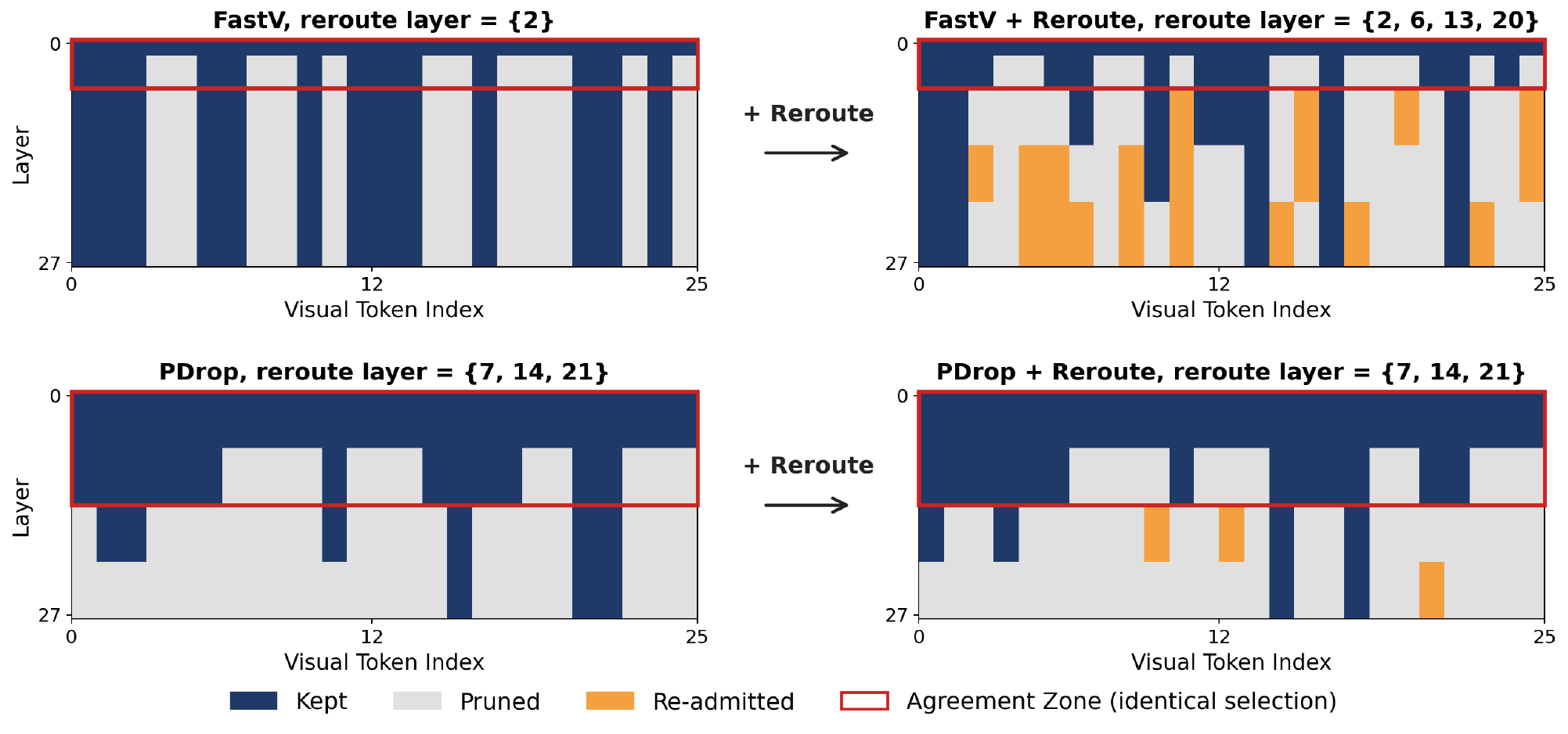}

    \vspace{-0.25em}
    \caption{
\textbf{Pruning as a degenerate routing policy.}
We compare FastV and PDrop with their Reroute variants on Qwen2.5-VL-7B.
Blue, gray, and orange denote selected, pruned/deferred, and re-admitted tokens.
The red box marks the shared selected-token prefix before pruning and reroute diverge: pruning keeps only the previous candidate set, whereas reroute restores the full visual-token candidate set.
}
    \label{fig:concept}
\vspace{-2mm}
\end{figure}

\vspace{-2mm}
\section{Method}
\label{sec:method}

\vspace{-2mm}
\subsection{Reroute: recoverable stage-wise routing.}
\label{sec:reroute}
\vspace{-2mm}

Reroute keeps the stage-wise ranking interface of pruning but replaces the post-selection action. At routing layer $\ell_i$, the same scorer and keep-ratio schedule yield $\mathcal{V}_i^{\mathrm{sel}}$ and $\mathcal{V}_i^{\mathrm{def}}$: selected tokens traverse the full decoder path, while deferred tokens bypass the current stage yet remain eligible for later selection. As summarized in Figure~\ref{fig:pipeline}, each stage ranks vision tokens via text-to-vision attention, routes $\mathcal{V}_i^{\mathrm{sel}}$ through Attn+FFN, and defers the rest until the next routing decision.

Let \(\operatorname{Block}_{\ell}\) denote the full decoder block at layer \(\ell\) (self-attention, FFN, residuals, and normalization). Reroute applies it only to the active subsequence \(H_{\mathcal{A}_i}^{\ell}\), scatters outputs to their original positions, and leaves deferred tokens untouched:
\[
h_j^{\ell+1} =
\begin{cases}
\operatorname{Block}_{\ell}(H_{\mathcal{A}_i}^{\ell})_j, & j \in \mathcal{A}_i, \\
h_j^\ell, & j \in \mathcal{V}_i^{\mathrm{def}}.
\end{cases}
\]
At the next routing layer, the full candidate set is restored by
$\mathcal{C}_{i+1}^{\mathrm{reroute}} = \mathcal{V}$, so a token deferred in one stage may be re-selected later if its score rises. Reroute thus converts irreversible deletion into recoverable deferral.

\vspace{-2mm}
\subsection{Pruning as a degenerate routing policy.}
\label{sec:pruning-as-degenerate-routing}
\vspace{-2mm}

The two methods differ only in the candidate-set transition:
\begin{equation}
\begin{aligned}
\mathcal{C}_{i+1}^{\mathrm{prune}} = \mathcal{V}_i^{\mathrm{sel}}
&& \text{and} &&
\mathcal{C}_{i+1}^{\mathrm{reroute}} = \mathcal{V}.
\end{aligned}
\end{equation}
Sharing the same scoring rule, routing layers, and stage budgets, pruning is the degenerate policy that assigns zero re-entry probability to deferred tokens, whereas Reroute keeps them valid for future stages. Figure~\ref{fig:concept} illustrates this: prior to an additional reroute decision the two policies can produce identical selections, but later stages may re-admit tokens that pruning has already discarded.

\vspace{-2mm}
\subsection{Matched-budget efficiency.}
\label{sec:matched-budget-efficiency}
\vspace{-2mm}

Reroute is designed for budget-matched comparison with the pruning baseline it augments. Under a shared schedule $\{(\ell_i,r_i)\}_{i=1}^S$, both methods activate the same number of vision tokens per stage. Token identities may diverge once re-entry is allowed. The active set is $\mathcal{A}_i = \mathcal{T} \cup \mathcal{V}_i^{\mathrm{sel}}$. With an optimized implementation, only $\mathcal{A}_i$ executes the decoder block at stage $i$, so Reroute targets the same theoretical TFLOPs and KV-cache class as the corresponding pruning schedule. The sole distinction is the post-selection action: pruning discards $\mathcal{V}_i^{\mathrm{def}}$, while Reroute retains these tokens in the sequence state for future routing. Practical latency still depends on gather/scatter kernels and KV-cache management.

\vspace{-2mm}
\subsection{Relation to conditional computation.}
\label{sec:relation-to-conditional-computation}
\vspace{-2mm}

Reroute can be viewed as a training-free application of Mixture-of-Depth (MoD)-style conditional computation to VLM visual-token pruning. Whereas standard MoD methods learn a router that decides which tokens execute a block and which bypass it, Reroute reuses the same conditional-computation pattern but inherits its routing signal, without training, from the text-to-vision attention scores already used by pruning.

This framing clarifies the design motivation. Visual-token pruning already poses a token-wise compute-allocation problem: selected tokens receive decoder computation while low-ranked ones are excluded. Rank-and-remove resolves the allocation by deletion. Reroute instead treats the excluded tokens as a bypass branch. They skip the stage but remain eligible for later selection. Reroute is therefore not a new general-purpose MoD architecture, but a training-free instantiation of the MoD principle in the MLLM visual-token pruning setting, with irreversible removal replaced by recoverable deferral.

\newcommand{\best}[1]{\textbf{#1}}
\newcommand{\second}[1]{\underline{#1}}
\newcommand{\na}{--}
\newcommand{\NuwaOfficial}{\textsuperscript{*}}

\definecolor{vanillarow}{HTML}{EFEFEF}
\definecolor{otherrow}{HTML}{EEF3FF}
\definecolor{baserow}{HTML}{FFF7E6}
\definecolor{routerow}{HTML}{EAF6EA}







\vspace{-2mm}
\section{Experiments}
\label{sec:experiments}
\vspace{-2mm}

We evaluate Reroute under matched visual-token budgets on grounding, general VQA, cross-backbone transfer, and efficiency.

\vspace{-2mm}
\subsection{Experimental setup}
\label{sec:exp_setup}

\vspace{-2mm}
\textbf{Benchmarks.}
We evaluate grounding on RefCOCO, RefCOCO+, and RefCOCOg~\cite{kazemzadeh2014referitgame, yu2016modeling} using bounding-box accuracy at IoU~0.5 (Acc@0.5). 
To assess general multimodal performance under visual-token reduction, we also report results on VQA benchmarks~\cite{liu2024mmbench, yue2024mmmu, li2023evaluating, lu2022learn, li2023seed}, including GQA~\cite{hudson2019gqa}, TextVQA~\cite{singh2019towards}, and MME~\cite{fu2023mme}, depending on model support and completed runs.

\vspace{-2mm}
\textbf{Models.}
We experiment with LLaVA-1.5-7B~\cite{liu2024improved} and Qwen-family models, including Qwen2.5-VL-7B~\cite{bai2025qwen25vltechnicalreport} and Qwen3.5-9B-Hybrid~\cite{qwen3.5}. LLaVA-1.5-7B is used for comparison with prior pruning methods, while Qwen models test cross-backbone transfer. Checkpoint sources and implementation formats are marked separately in the tables.

\vspace{-2mm}
\textbf{Baselines.}
We compare against the full model, FastV~\cite{chen2024image}, PDrop~\cite{xing2024pyramiddrop}, and their budget-matched Reroute variants~\cite{zhang2025beyond, zhang2024sparsevlm,yang2025visionzip,endo2025feather,huang2026n}. We also consider Nüwa~\cite{huang2026n} as a strong recent baseline: using its official codebase, we augment its FastV-style decoder pruning with Reroute. 



\definecolor{vanillarow}{HTML}{EFEFEF}
\definecolor{otherrow}{HTML}{F2F6FF}
\definecolor{baserow}{HTML}{FFF7E6}
\definecolor{routerow}{HTML}{EAF6EA}

\newcolumntype{M}[1]{>{\raggedright\arraybackslash}m{#1}} 
\newcolumntype{S}[1]{>{\centering\arraybackslash}m{#1}}  
\newcolumntype{Z}[1]{>{\centering\arraybackslash}m{#1}}  

\newcommand{\src}[1]{{\fontsize{5.6}{6.0}\selectfont #1}}

\newcommand{\Arowstrut}{\rule{0pt}{7.55pt}}
\newcommand{\Browstrut}{\rule{0pt}{8.00pt}}

\newcommand{\ApreTokB}{\addlinespace[0.00pt]}
\newcommand{\ApreTokC}{\addlinespace[1.05pt]}
\newcommand{\ApreTokD}{\addlinespace[0.00pt]}

\newcommand{\BpreTokB}{\addlinespace[0.00pt]}
\newcommand{\BpreTokC}{\addlinespace[0.00pt]}
\newcommand{\BpreTokD}{\addlinespace[1.35pt]}

\begin{table*}[t]
\centering
\caption{
\textbf{Visual grounding performance on RefCOCO-series benchmarks using LLaVA-1.5-7B.}
We compare visual-token reduction methods and their Reroute variants under matched average-token budgets.
Panel (a) reports the original LLaVA-format results, with \NuwaOfficial{} marking N\"uwa results reproduced in the official N\"uwa codebase; panel (b) reports HuggingFace-format reproduced results under the same LLaVA-1.5-7B backbone.
When shown, A/B denote testA/testB splits.
Avg. Ratio is normalized by the corresponding full-token baseline within each panel.
Best and second-best results within each average-token block are shown in \best{bold} and \second{underlined}, respectively, and unavailable results are denoted by ``--''.
}
\label{tab:llava15_refcoco_combined_side_by_side}
\vspace{-2mm}
\scriptsize
\setlength{\tabcolsep}{0.95pt}
\renewcommand{\arraystretch}{0.78}
\setlength{\aboverulesep}{0.18ex}
\setlength{\belowrulesep}{0.18ex}
\setlength{\cmidrulesep}{0.10ex}

\noindent
\begin{minipage}[t]{0.492\textwidth}
\centering
\textbf{(a) Original LLaVA format results.}
\vspace{0.10em}

\resizebox{\linewidth}{!}{%
\begin{tabular}{@{}M{6.9em}S{3.3em}*{7}{Z{2.15em}}Z{3.35em}@{}}
\toprule
\textbf{Method} & \textbf{Src.}
& \multicolumn{2}{c}{\textbf{RefCOCO}}
& \multicolumn{3}{c}{\textbf{RefCOCO+}}
& \multicolumn{2}{c}{\textbf{RefCOCOg}}
& \textbf{Avg.} \\
\cmidrule(lr){3-4}
\cmidrule(lr){5-7}
\cmidrule(lr){8-9}
& & \textbf{val} & \textbf{test}
& \textbf{val} & \textbf{A} & \textbf{B}
& \textbf{val} & \textbf{test}
& \textbf{Ratio} \\
\midrule

\multicolumn{10}{c}{\textit{Average Token 576}} \\
\midrule

\rowcolor{vanillarow}
\Arowstrut Vanilla\NuwaOfficial
& \src{CVPR'24}
& \best{56.2} & \best{58.0}
& \best{50.0} & \best{59.3} & \best{38.8}
& \best{48.9} & \best{48.4}
& \best{100.0\%} \\

\ApreTokB
\midrule
\multicolumn{10}{c}{\textit{Average Token 192} $\downarrow 66.7\%$} \\
\midrule

\rowcolor{otherrow}
\Arowstrut FEATHER
& \src{ICCV'25}
& \na & 27.7
& \na & 2.5 & \na
& \na & 27.2
& 36.1\% \\

\midrule

\rowcolor{baserow}
\Arowstrut N\"uwa\NuwaOfficial
& \src{ICLR'26}
& \best{59.1} & \best{60.1}
& \best{50.5} & \second{60.9} & \best{40.2}
& \best{51.8} & \best{50.1}
& \best{103.7\%} \\

\rowcolor{routerow}
\Arowstrut N\"uwa+Reroute\NuwaOfficial
& \src{Ours}
& \second{58.6} & \second{59.6}
& \second{50.4} & \best{61.3} & \second{39.8}
& \second{50.9} & \second{49.9}
& \second{103.0\%} \\

\ApreTokC
\midrule
\multicolumn{10}{c}{\textit{Average Token 128} $\downarrow 77.8\%$} \\
\midrule

\rowcolor{otherrow}
\Arowstrut FastV
& \src{ECCV'24}
& 10.1 & 10.3
& 8.2 & 8.5 & 9.8
& 9.1 & 8.9
& 18.4\% \\

\rowcolor{otherrow}
\Arowstrut SparseVLM
& \src{ICML'25}
& 6.2 & 6.3
& 9.9 & 5.8 & 4.2
& 6.5 & 6.4
& 12.6\% \\

\rowcolor{otherrow}
\Arowstrut VisionZip
& \src{CVPR'25}
& 4.1 & 4.5
& 3.9 & 4.1 & 4.9
& 3.5 & 3.5
& 8.1\% \\

\midrule

\rowcolor{baserow}
\Arowstrut N\"uwa\NuwaOfficial
& \src{ICLR'26}
& \best{52.3} & \best{53.5}
& \second{43.7} & \second{53.2} & \best{35.6}
& \best{45.0} & \best{43.5}
& \best{90.9\%} \\

\rowcolor{routerow}
\Arowstrut N\"uwa+Reroute\NuwaOfficial
& \src{Ours}
& \second{51.8} & \second{53.0}
& \best{44.3} & \best{55.5} & \second{35.0}
& \second{44.3} & \second{43.1}
& \second{90.3\%} \\

\ApreTokD
\midrule
\multicolumn{10}{c}{\textit{Average Token 64} $\downarrow 88.9\%$} \\
\midrule

\rowcolor{otherrow}
\Arowstrut FastV
& \src{ECCV'24}
& 2.0 & 2.7
& 2.4 & 1.2 & 1.0
& 2.0 & 2.2
& 3.8\% \\

\rowcolor{otherrow}
\Arowstrut SparseVLM
& \src{ICML'25}
& 1.0 & 1.0
& 1.0 & 1.0 & 1.3
& 0.7 & 0.6
& 1.9\% \\

\rowcolor{otherrow}
\Arowstrut VisionZip
& \src{CVPR'25}
& 3.8 & 4.0
& 3.5 & 3.7 & 3.9
& 3.2 & 3.4
& 7.2\% \\

\midrule

\rowcolor{baserow}
\Arowstrut N\"uwa\NuwaOfficial
& \src{ICLR'26}
& \best{32.8} & \second{33.2}
& \second{26.5} & \second{34.3} & \second{20.5}
& \second{25.6} & \second{24.5}
& \second{54.6\%} \\

\rowcolor{routerow}
\Arowstrut N\"uwa+Reroute\NuwaOfficial
& \src{Ours}
& \best{32.8} & \best{33.7}
& \best{27.2} & \best{35.3} & \best{21.1}
& \best{26.3} & \best{25.2}
& \best{55.8\%} \\

\bottomrule
\end{tabular}%
}
\end{minipage}%
\hfill%
\begin{minipage}[t]{0.492\textwidth}
\centering
\textbf{(b) HuggingFace-format reproduced results.}
\vspace{0.10em}

\resizebox{\linewidth}{!}{%
\begin{tabular}{@{}M{6.7em}S{3.3em}*{8}{Z{2.05em}}Z{3.35em}@{}}
\toprule
\textbf{Method} & \textbf{Src.}
& \multicolumn{3}{c}{\textbf{RefCOCO}}
& \multicolumn{3}{c}{\textbf{RefCOCO+}}
& \multicolumn{2}{c}{\textbf{RefCOCOg}}
& \textbf{Avg.} \\
\cmidrule(lr){3-5}
\cmidrule(lr){6-8}
\cmidrule(lr){9-10}
& & \textbf{val} & \textbf{A} & \textbf{B}
& \textbf{val} & \textbf{A} & \textbf{B}
& \textbf{val} & \textbf{test}
& \textbf{Ratio} \\
\midrule

\multicolumn{11}{c}{\textit{Average Token 576}} \\
\midrule

\rowcolor{vanillarow}
\Browstrut Vanilla (HF)
& \src{CVPR'24}
& \best{54.0} & \best{59.4} & \best{48.0}
& \best{46.5} & \best{54.1} & \best{39.1}
& \best{46.9} & \best{47.6}
& \best{100.0\%} \\

\BpreTokB
\midrule
\multicolumn{11}{c}{\textit{Average Token 192} $\downarrow 66.7\%$} \\
\midrule

\rowcolor{baserow}
\Browstrut FastV
& \src{ECCV'24}
& 25.3 & 27.8 & 22.3
& 21.0 & 23.6 & 18.0
& 21.5 & 20.5
& 45.5\% \\

\rowcolor{routerow}
\Browstrut FastV+Reroute
& \src{Ours}
& 32.7 & 37.7 & 26.8
& 26.7 & 32.3 & 21.3
& 28.7 & 29.1
& 59.2\% \\

\rowcolor{baserow}
\Browstrut PDrop
& \src{CVPR'25}
& \second{39.8} & \second{45.1} & \second{33.4}
& \second{33.5} & \second{39.5} & \second{25.9}
& \second{33.2} & \second{32.8}
& \second{71.3\%} \\

\rowcolor{routerow}
\Browstrut PDrop+Reroute
& \src{Ours}
& \best{45.3} & \best{49.5} & \best{39.0}
& \best{37.9} & \best{44.3} & \best{31.1}
& \best{39.7} & \best{39.2}
& \best{82.3\%} \\

\BpreTokC
\midrule
\multicolumn{11}{c}{\textit{Average Token 128} $\downarrow 77.8\%$} \\
\midrule

\rowcolor{baserow}
\Browstrut FastV
& \src{ECCV'24}
& 11.0 & 12.7 & 9.1
& 8.9 & 10.3 & 8.0
& 9.3 & 9.0
& 19.8\% \\

\rowcolor{routerow}
\Browstrut FastV+Reroute
& \src{Ours}
& 17.4 & 20.7 & 13.9
& 14.5 & 17.5 & 11.6
& 15.8 & 15.1
& 31.8\% \\

\rowcolor{baserow}
\Browstrut PDrop
& \src{CVPR'25}
& \second{30.7} & \second{36.5} & \second{23.7}
& \second{25.2} & \second{30.5} & \second{18.3}
& \second{24.4} & \second{23.9}
& \second{53.4\%} \\

\rowcolor{routerow}
\Browstrut PDrop+Reroute
& \src{Ours}
& \best{34.1} & \best{39.6} & \best{27.2}
& \best{27.6} & \best{33.2} & \best{22.0}
& \best{29.8} & \best{29.7}
& \best{61.2\%} \\

\BpreTokD
\midrule
\multicolumn{11}{c}{\textit{Average Token 64} $\downarrow 88.9\%$} \\
\midrule

\rowcolor{baserow}
\Browstrut FastV
& \src{ECCV'24}
& 0.3 & 0.2 & 0.6
& 0.3 & 0.1 & 0.5
& 0.1 & 0.1
& 0.6\% \\

\rowcolor{routerow}
\Browstrut FastV+Reroute
& \src{Ours}
& 0.4 & 0.2 & 0.4
& 0.3 & 0.1 & 0.7
& 0.1 & 0.1
& 0.6\% \\

\rowcolor{baserow}
\Browstrut PDrop
& \src{CVPR'25}
& \second{12.8} & \second{14.8} & \second{10.2}
& \second{10.1} & \second{12.1} & \second{8.0}
& \second{10.3} & \second{10.1}
& \second{22.2\%} \\

\rowcolor{routerow}
\Browstrut PDrop+Reroute
& \src{Ours}
& \best{18.9} & \best{23.3} & \best{14.5}
& \best{15.4} & \best{19.9} & \best{11.5}
& \best{16.5} & \best{15.6}
& \best{34.0\%} \\

\bottomrule
\end{tabular}%
}
\end{minipage}

\vspace{0.35em}
\begin{minipage}{0.96\textwidth}
\centering
\scriptsize
\setlength{\fboxsep}{1.2pt}

\colorbox{vanillarow}{\strut Vanilla baseline}\quad
\colorbox{otherrow}{\strut Other methods}\quad
\colorbox{baserow}{\strut Base methods}\quad
\colorbox{routerow}{\strut Reroute variants}.
\end{minipage}
\end{table*}

\vspace{-2mm}
\subsection{Main results}

\vspace{-2mm}
\paragraph{Visual grounding on LLaVA-1.5.}
\label{sec:main-results-llava-grounding}
Table~\ref{tab:llava15_refcoco_combined_side_by_side} reports RefCOCO-series results on LLaVA-1.5 under two checkpoint formats. Panel~(a) follows the original LLaVA-1.5 format, combining paper-reported baselines with rows reproduced in the official Nüwa codebase ($^*$). Panel~(b) uses the HuggingFace-format checkpoint and reports fully reproduced results from our unified pipeline. Results are grouped by average visual-token budget, and average ratios are normalized by the corresponding full-token baseline.

In the HuggingFace-format setting, Reroute consistently improves matched pruning baselines. At 192 average tokens, it raises the average ratio from 45.5\% to 59.2\% for FastV and from 71.3\% to 82.3\% for PDrop. At 128 tokens, the gains remain 12.0 and 7.8 points, respectively. At the extreme 64-token budget, one-shot FastV leaves little room for recovery, whereas PDrop+Reroute still improves from 22.2\% to 34.0\%, indicating that repeated re-entry opportunities are important for effective rerouting.

In the original-format setting, Nüwa is already a strong grounding-oriented baseline. Reroute matches it at moderate budgets and further improves the average ratio at 64 tokens from 54.6\% to 55.8\%, showing that recoverable deferral remains beneficial even on top of a stronger pipeline.

Overall, these results support our view of Reroute as a change in the post-selection action rather than a new scoring heuristic: replacing irreversible deletion with recoverable deferral improves FastV/PDrop-style pruning, especially under stage-wise schedules, while remaining compatible with stronger grounding-aware pipelines.



\definecolor{vanillarow}{HTML}{EFEFEF}
\definecolor{baserow}{HTML}{FFF7E6}
\definecolor{routerow}{HTML}{EAF6EA}

\newcolumntype{M}[1]{>{\raggedright\arraybackslash}m{#1}}
\newcolumntype{S}[1]{>{\centering\arraybackslash}m{#1}}
\newcolumntype{Z}[1]{>{\centering\arraybackslash}m{#1}}


\begin{table*}[t]
\centering
\caption{\textbf{RefCOCO-series visual grounding with Qwen-family VLMs.}
FastV, PDrop, and their Reroute variants are compared under matched average-token budgets.
Panel (a) uses Qwen2.5-VL-7B with decoder-side retained visual tokens; panel (b) uses Qwen3.5-9B-Hybrid with gated-attention routed visual tokens.
A/B denote testA/testB.
Avg. Ratio is normalized to the corresponding vanilla baseline.
Best and second-best results within each budget block are bolded and underlined.}
\label{tab:qwen_refcoco_combined}
\vspace{-2mm}
\scriptsize
\setlength{\tabcolsep}{0.95pt}
\renewcommand{\arraystretch}{0.78}
\setlength{\aboverulesep}{0.18ex}
\setlength{\belowrulesep}{0.18ex}
\setlength{\cmidrulesep}{0.10ex}

\noindent
\begin{minipage}[t]{0.492\textwidth}
\centering
\textbf{(a) Qwen2.5-VL-7B.}
\vspace{0.10em}

\resizebox{\linewidth}{!}{%
\begin{tabular}{@{}M{6.8em}S{3.3em}*{8}{Z{2.05em}}Z{3.25em}@{}}
\toprule
\textbf{Method} & \textbf{Src.}
& \multicolumn{3}{c}{\textbf{RefCOCO}}
& \multicolumn{3}{c}{\textbf{RefCOCO+}}
& \multicolumn{2}{c}{\textbf{RefCOCOg}}
& \textbf{Avg.} \\
\cmidrule(lr){3-5}
\cmidrule(lr){6-8}
\cmidrule(lr){9-10}
& & \textbf{val} & \textbf{A} & \textbf{B}
& \textbf{val} & \textbf{A} & \textbf{B}
& \textbf{val} & \textbf{test}
& \textbf{Ratio} \\
\midrule

\multicolumn{11}{c}{\textit{Average Token 100\%}} \\
\midrule
\rowcolor{vanillarow}
Vanilla
& \src{Qwen'25}
& \best{87.7} & \best{89.8} & \best{82.3}
& \best{80.5} & \best{85.9} & \best{72.8}
& \best{82.7} & \best{83.3}
& \best{100.0\%} \\

\midrule
\multicolumn{11}{c}{\textit{Average Token Reduction} $\downarrow 66.7\%$} \\
\midrule
\rowcolor{baserow}
FastV
& \src{ECCV'24}
& 46.4 & 48.1 & 44.4
& 40.5 & 42.6 & 39.3
& 48.4 & 47.2
& 53.7\% \\

\rowcolor{routerow}
FastV+Reroute
& \src{Ours}
& 53.4 & 57.8 & 49.4
& 45.0 & 50.9 & 40.6
& 52.8 & 52.1
& 60.3\% \\

\addlinespace[1.0pt]

\rowcolor{baserow}
PDrop
& \src{CVPR'25}
& \second{55.9} & \second{59.2} & \second{52.0}
& \second{48.4} & \second{53.0} & \second{43.6}
& \second{53.6} & \second{52.9}
& \second{62.9\%} \\

\rowcolor{routerow}
PDrop+Reroute
& \src{Ours}
& \best{57.9} & \best{63.0} & \best{53.6}
& \best{49.9} & \best{55.7} & \best{44.8}
& \best{54.7} & \best{54.0}
& \best{65.1\%} \\

\midrule
\multicolumn{11}{c}{\textit{Average Token Reduction} $\downarrow 77.8\%$} \\
\midrule
\rowcolor{baserow}
FastV
& \src{ECCV'24}
& 19.2 & 21.5 & 17.8
& 16.7 & 18.4 & 14.6
& 19.1 & 18.6
& 21.9\% \\

\rowcolor{routerow}
FastV+Reroute
& \src{Ours}
& \second{26.8} & \second{31.1} & 22.9
& \second{21.7} & 26.0 & 18.0
& \second{25.3} & \second{24.1}
& 29.3\% \\

\addlinespace[1.0pt]

\rowcolor{baserow}
PDrop
& \src{CVPR'25}
& 25.9 & 30.9 & \second{24.2}
& 21.2 & \second{26.7} & \second{19.2}
& 24.1 & 23.7
& \second{29.3\%} \\

\rowcolor{routerow}
PDrop+Reroute
& \src{Ours}
& \best{33.2} & \best{37.6} & \best{31.5}
& \best{27.7} & \best{33.5} & \best{25.1}
& \best{31.0} & \best{29.7}
& \best{37.4\%} \\

\midrule
\multicolumn{11}{c}{\textit{Average Token Reduction} $\downarrow 88.9\%$} \\
\midrule
\rowcolor{baserow}
FastV
& \src{ECCV'24}
& 1.6 & 1.6 & 2.2
& 1.7 & 1.7 & 2.6
& 2.0 & 2.3
& 2.4\% \\

\rowcolor{routerow}
FastV+Reroute
& \src{Ours}
& 6.2 & 5.5 & 5.9
& 4.8 & 4.2 & 4.8
& 6.7 & 6.5
& 6.7\% \\

\addlinespace[1.0pt]

\rowcolor{baserow}
PDrop
& \src{CVPR'25}
& \second{10.2} & \second{10.9} & \second{9.0}
& \second{7.9} & \second{9.8} & \second{6.7}
& \second{9.9} & \second{9.5}
& \second{11.1\%} \\

\rowcolor{routerow}
PDrop+Reroute
& \src{Ours}
& \best{16.8} & \best{19.2} & \best{15.2}
& \best{13.9} & \best{16.9} & \best{12.2}
& \best{15.0} & \best{14.9}
& \best{18.6\%} \\

\bottomrule
\end{tabular}%
}
\end{minipage}%
\hfill%
\begin{minipage}[t]{0.492\textwidth}
\centering
\textbf{(b) Qwen3.5-9B-Hybrid.}
\vspace{0.10em}

\resizebox{\linewidth}{!}{%
\begin{tabular}{@{}M{6.8em}S{3.3em}*{8}{Z{2.05em}}Z{3.25em}@{}}
\toprule
\textbf{Method} & \textbf{Src.}
& \multicolumn{3}{c}{\textbf{RefCOCO}}
& \multicolumn{3}{c}{\textbf{RefCOCO+}}
& \multicolumn{2}{c}{\textbf{RefCOCOg}}
& \textbf{Avg.} \\
\cmidrule(lr){3-5}
\cmidrule(lr){6-8}
\cmidrule(lr){9-10}
& & \textbf{val} & \textbf{A} & \textbf{B}
& \textbf{val} & \textbf{A} & \textbf{B}
& \textbf{val} & \textbf{test}
& \textbf{Ratio} \\
\midrule

\multicolumn{11}{c}{\textit{Average Gated Attention Token 100\%}} \\
\midrule
\rowcolor{vanillarow}
Vanilla
& \src{Qwen'25}
& \best{90.4} & \best{90.7} & \best{85.5}
& \best{80.0} & \best{84.3} & \best{75.7}
& \best{87.3} & \best{86.8}
& \best{100.0\%} \\

\midrule
\multicolumn{11}{c}{\textit{Average Gated Attention Token Reduction} $\downarrow 66.7\%$} \\
\midrule
\rowcolor{baserow}
FastV
& \src{ECCV'24}
& 28.7 & 32.5 & 25.6
& 25.2 & 28.3 & 20.3
& 26.3 & 25.6
& 31.1\% \\

\rowcolor{routerow}
FastV+Reroute
& \src{Ours}
& \best{70.8} & \second{75.1} & \best{64.8}
& \second{60.7} & \second{67.0} & \second{53.9}
& \second{68.5} & \second{68.0}
& \second{77.5\%} \\

\addlinespace[1.0pt]

\rowcolor{baserow}
PDrop
& \src{CVPR'25}
& 36.4 & 40.4 & 32.4
& 32.7 & 36.1 & 27.4
& 33.7 & 33.4
& 40.0\% \\

\rowcolor{routerow}
PDrop+Reroute
& \src{Ours}
& \second{70.0} & \best{76.1} & \second{62.4}
& \best{61.1} & \best{68.6} & \best{54.2}
& \best{69.2} & \best{68.3}
& \best{77.7\%} \\

\midrule
\multicolumn{11}{c}{\textit{Average Gated Attention Token Reduction} $\downarrow 77.8\%$} \\
\midrule
\rowcolor{baserow}
FastV
& \src{ECCV'24}
& 18.0 & 22.2 & 15.9
& 14.5 & 18.3 & 12.3
& 16.6 & 15.9
& 19.5\% \\

\rowcolor{routerow}
FastV+Reroute
& \src{Ours}
& \second{51.7} & \second{58.4} & \second{44.6}
& \second{42.3} & \second{49.7} & \second{35.7}
& \second{50.2} & \second{49.6}
& \second{55.9\%} \\

\addlinespace[1.0pt]

\rowcolor{baserow}
PDrop
& \src{CVPR'25}
& 23.9 & 28.6 & 20.4
& 20.4 & 25.0 & 16.7
& 20.8 & 20.4
& 25.8\% \\

\rowcolor{routerow}
PDrop+Reroute
& \src{Ours}
& \best{54.9} & \best{62.7} & \best{47.0}
& \best{46.4} & \best{53.5} & \best{38.0}
& \best{52.9} & \best{52.2}
& \best{59.7\%} \\

\midrule
\multicolumn{11}{c}{\textit{Average Gated Attention Token Reduction} $\downarrow 88.9\%$} \\
\midrule
\rowcolor{baserow}
FastV
& \src{ECCV'24}
& 12.4 & 14.7 & 12.3
& 6.8 & 6.8 & 5.8
& 8.3 & 8.1
& 10.9\% \\

\rowcolor{routerow}
FastV+Reroute
& \src{Ours}
& 14.2 & 15.0 & 13.2
& 8.8 & 9.4 & 7.6
& 11.9 & 11.6
& 13.4\% \\

\addlinespace[1.0pt]

\rowcolor{baserow}
PDrop
& \src{CVPR'25}
& \second{19.5} & \second{23.8} & \second{15.7}
& \second{14.2} & \second{17.8} & \second{10.0}
& \second{15.8} & \second{15.5}
& \second{19.3\%} \\

\rowcolor{routerow}
PDrop+Reroute
& \src{Ours}
& \best{35.0} & \best{41.1} & \best{27.8}
& \best{27.8} & \best{34.8} & \best{22.2}
& \best{33.7} & \best{33.8}
& \best{37.4\%} \\

\bottomrule
\end{tabular}%
}
\end{minipage}

\vspace{0.35em}
\begin{minipage}{0.96\textwidth}
\centering
\scriptsize
\setlength{\fboxsep}{1.2pt}

\colorbox{vanillarow}{\strut Vanilla baseline}\quad
\colorbox{baserow}{\strut Base methods}\quad
\colorbox{routerow}{\strut Reroute variants}.
\end{minipage}

\end{table*}

\vspace{-2mm}
\paragraph{General VQA results on LLaVA-1.5.}
\label{sec:general-vqa-llava15}
Table~\ref{tab:llava15_vqa_hf_format} reports general VQA results on the HuggingFace-format LLaVA-1.5 checkpoint. These benchmarks are used as an auxiliary evaluation: our main hypothesis concerns recoverable visual evidence for grounding under aggressive token reduction, while VQA measures whether the same routing mechanism preserves general multimodal capability.

Across the 192-, 128-, and 64-token budgets, Reroute remains comparable to or better than its matched FastV and PDrop counterparts. This indicates that the grounding improvements reported above are not obtained by sacrificing general VQA performance. Instead, recoverable routing preserves broad multimodal behavior while providing larger benefits on grounding-sensitive tasks, where irreversible removal is more likely to discard visual evidence needed by later decoder layers.


\providecommand{\na}{--}
\providecommand{\best}[1]{\textbf{#1}}
\providecommand{\second}[1]{\underline{#1}}
\providecommand{\effdown}[1]{\textcolor{green!55!black}{$\downarrow$ #1\%}}

\definecolor{vanillarow}{HTML}{EFEFEF}
\definecolor{baserow}{HTML}{FFF7E6}
\definecolor{routerow}{HTML}{EAF6EA}

\begin{table*}[t]
\centering

\begin{minipage}[t]{0.56\textwidth}
\centering
\vspace{-2mm}
\caption{
\textbf{VQA performance on LLaVA-1.5-7B under the HuggingFace model format.}
Bold and underline denote best and second-best results within each average-token block.
}
\label{tab:llava15_vqa_hf_format}
\vspace{-2mm}
\scriptsize
\setlength{\tabcolsep}{2.3pt}
\renewcommand{\arraystretch}{0.90}
\setlength{\aboverulesep}{0.20ex}
\setlength{\belowrulesep}{0.20ex}
\setlength{\cmidrulesep}{0.08ex}

\resizebox{\linewidth}{!}{%
\begin{tabular}{@{}lccccccccc@{}}
\toprule
\textbf{Method}
& \textbf{GQA} & \textbf{MMB} & \textbf{MMMU} & \textbf{MME}
& \textbf{TextVQA} & \textbf{POPE} & \textbf{SQA}
& \textbf{SEED} & \textbf{Avg. Ratio} \\
\midrule

\multicolumn{10}{c}{\textit{Average Token 576}} \\
\midrule
\rowcolor{vanillarow}
Vanilla
& \best{60.5} & \best{62.4} & \best{34.4} & \best{1782}
& \best{57.5} & \best{85.9} & \best{66.1}
& \best{65.5} & \best{100.0\%} \\

\midrule
\multicolumn{10}{c}{\textit{Average Token 192} $\downarrow 66.7\%$} \\
\midrule
\rowcolor{baserow}
FastV
& 55.0 & 58.6 & 34.2 & 1637
& 53.7 & 79.6 & 66.4
& 58.2 & 93.9\% \\

\rowcolor{routerow}
FastV+Reroute
& 57.4 & 60.7 & 34.2 & 1672
& \second{56.7} & 81.0 & \second{67.1}
& 60.4 & 96.5\% \\

\rowcolor{baserow}
PDrop
& \best{59.2} & \best{61.4} & \best{34.7} & \best{1776}
& 46.9 & \best{84.4} & 65.9
& \best{63.4} & \second{96.6\%} \\

\rowcolor{routerow}
PDrop+Reroute
& \second{58.9} & \second{61.3} & \second{34.4} & \second{1775}
& \best{57.4} & \second{84.2} & 66.1
& \second{62.6} & \best{98.6\%} \\

\midrule
\multicolumn{10}{c}{\textit{Average Token 128} $\downarrow 77.8\%$} \\
\midrule
\rowcolor{baserow}
FastV
& 50.5 & 53.7 & 34.0 & 1451
& 50.4 & 73.0 & 65.5
& 54.3 & 88.0\% \\

\rowcolor{routerow}
FastV+Reroute
& 54.8 & 57.0 & \second{34.3} & 1557
& 53.6 & 75.1 & \best{68.4}
& 55.4 & 92.2\% \\

\rowcolor{baserow}
PDrop
& \second{57.0} & \second{59.5} & 34.2 & \second{1728}
& \second{56.2} & \best{82.1} & \second{66.8}
& \second{59.2} & \second{96.3\%} \\

\rowcolor{routerow}
PDrop+Reroute
& \best{57.3} & \best{59.8} & \best{34.6} & \best{1733}
& \best{56.5} & \second{80.9} & 66.6
& \best{60.8} & \best{96.8\%} \\

\midrule
\multicolumn{10}{c}{\textit{Average Token 64} $\downarrow 88.9\%$} \\
\midrule
\rowcolor{baserow}
FastV
& 40.0 & 22.4 & 32.1 & 950
& 41.8 & 51.1 & 65.1
& 36.0 & 66.8\% \\

\rowcolor{routerow}
FastV+Reroute
& 44.3 & 23.3 & 32.2 & 954
& 43.5 & 51.4 & 65.3
& 39.8 & 69.1\% \\

\rowcolor{baserow}
PDrop
& \second{52.8} & \second{55.1} & \second{33.1} & \second{1486}
& \second{51.6} & \best{74.9} & \second{68.2}
& \second{53.9} & \second{89.7\%} \\

\rowcolor{routerow}
PDrop+Reroute
& \best{54.4} & \best{57.4} & \best{33.8} & \best{1542}
& \best{52.5} & \second{74.8} & \best{69.2}
& \best{54.8} & \best{91.7\%} \\

\bottomrule
\end{tabular}%
}

\vspace{0.25em}
{\scriptsize
\setlength{\fboxsep}{0.9pt}
\colorbox{vanillarow}{\strut Vanilla baseline}\quad
\colorbox{baserow}{\strut Base methods}\quad
\colorbox{routerow}{\strut Reroute variants}
}

\end{minipage}
\hfill
\begin{minipage}[t]{0.42\textwidth}
\centering
\vspace{-2mm}
\caption{
\textbf{Efficiency comparison on LLaVA-1.5-7B.}
TFLOPs and KV-cache memory are normalized against the vanilla full-token baseline.
}
\label{tab:efficiency_comparison}
\vspace{-2mm}
\scriptsize
\setlength{\tabcolsep}{3.0pt}
\renewcommand{\arraystretch}{0.92}
\setlength{\aboverulesep}{0.22ex}
\setlength{\belowrulesep}{0.22ex}
\setlength{\cmidrulesep}{0.10ex}

\resizebox{\linewidth}{!}{%
\begin{tabular}{@{}l c cc cc@{}}
\toprule
\textbf{Method}
& \textbf{Avg.}
& \multicolumn{2}{c}{\textbf{TFLOPs}}
& \multicolumn{2}{c}{\textbf{KV Cache}} \\
\cmidrule(lr){3-4}
\cmidrule(l){5-6}
& \textbf{Tok.}
& \textbf{Val.} & \textbf{/Van.}
& \textbf{MB} & \textbf{/Van.} \\
\midrule

\rowcolor{vanillarow}
Vanilla
& 576
& 8.67 & \na
& 297.5 & \na \\

\midrule
\rowcolor{baserow}
FastV
& 306
& 4.89 & \effdown{44}
& 162.5 & \effdown{45} \\

\rowcolor{routerow}
FastV+Reroute
& 306
& 5.02 & \effdown{45}
& 162.5 & \effdown{45} \\

\midrule
\rowcolor{baserow}
FastV
& 128
& 2.45 & \effdown{72}
& 73.2 & \effdown{75} \\

\rowcolor{routerow}
FastV+Reroute
& 128
& 2.58 & \effdown{70}
& 73.2 & \effdown{75} \\

\rowcolor{baserow}
PDrop
& 128
& 2.47 & \effdown{72}
& 73.3 & \effdown{75} \\

\rowcolor{routerow}
PDrop+Reroute
& 128
& 2.56 & \effdown{75}
& 73.2 & \effdown{75} \\

\midrule
\rowcolor{baserow}
FastV
& 64
& 1.60 & \effdown{82}
& 41.5 & \effdown{86} \\

\rowcolor{routerow}
FastV+Reroute
& 64
& 1.73 & \effdown{80}
& 41.5 & \effdown{86} \\

\rowcolor{baserow}
PDrop
& 64
& 1.83 & \effdown{79}
& 49.7 & \effdown{83} \\

\rowcolor{routerow}
PDrop+Reroute
& 64
& 1.99 & \effdown{77}
& 51.6 & \effdown{83} \\

\bottomrule
\end{tabular}%
}

\vspace{0.25em}
{\scriptsize
\setlength{\fboxsep}{0.9pt}
\colorbox{vanillarow}{\strut Vanilla}\quad
\colorbox{baserow}{\strut Base}\quad
\colorbox{routerow}{\strut Reroute}
}

\end{minipage}

\vspace{-2mm}
\end{table*}
\begin{figure*}[t]
  \centering
  \vspace{-3mm}
  \captionsetup[subfigure]{font=small,skip=2pt}
  \captionsetup{font=small,skip=3pt}
  \begin{subfigure}[t]{0.492\textwidth}
    \centering
    \includegraphics[
      width=\linewidth,
      trim=2pt 2pt 2pt 2pt,
      clip
    ]{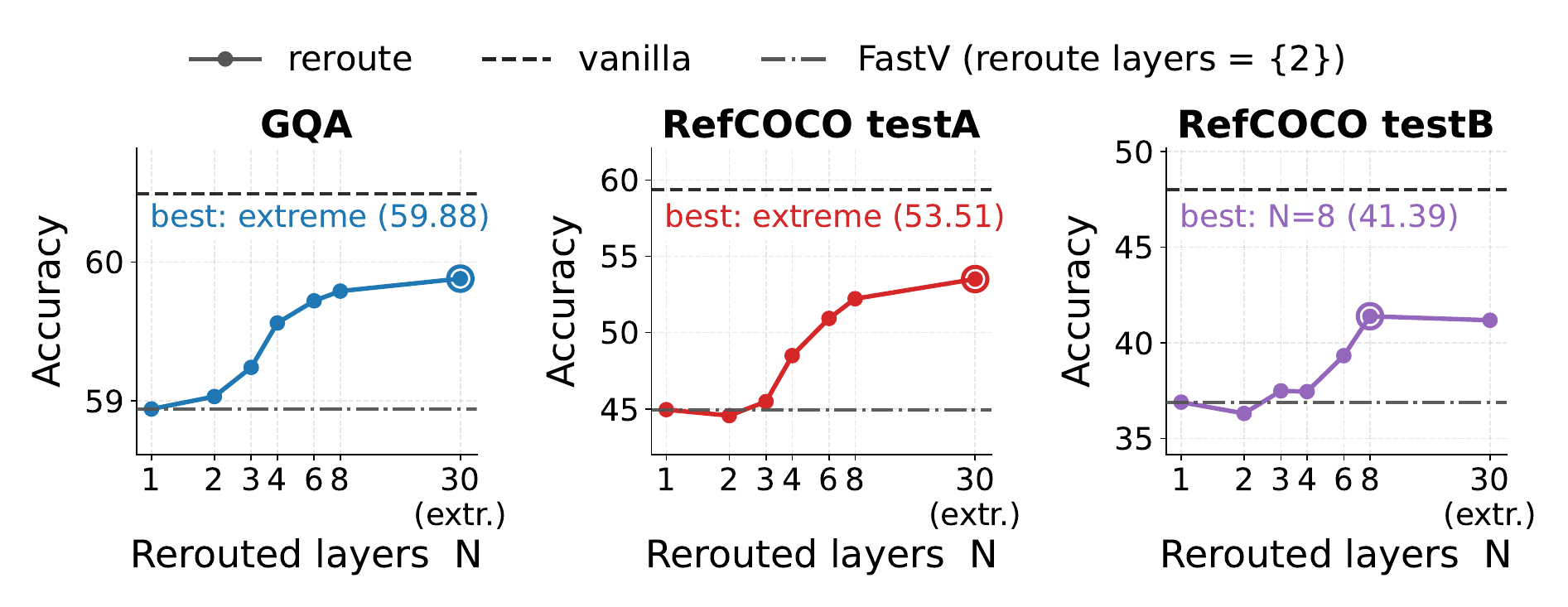}
    \caption{Frequency sweep.}
    \label{fig:ablation-freq}
  \end{subfigure}
  \hfill
  \begin{subfigure}[t]{0.492\textwidth}
    \centering
    \includegraphics[
      width=\linewidth,
      trim=2pt 2pt 2pt 2pt,
      clip
    ]{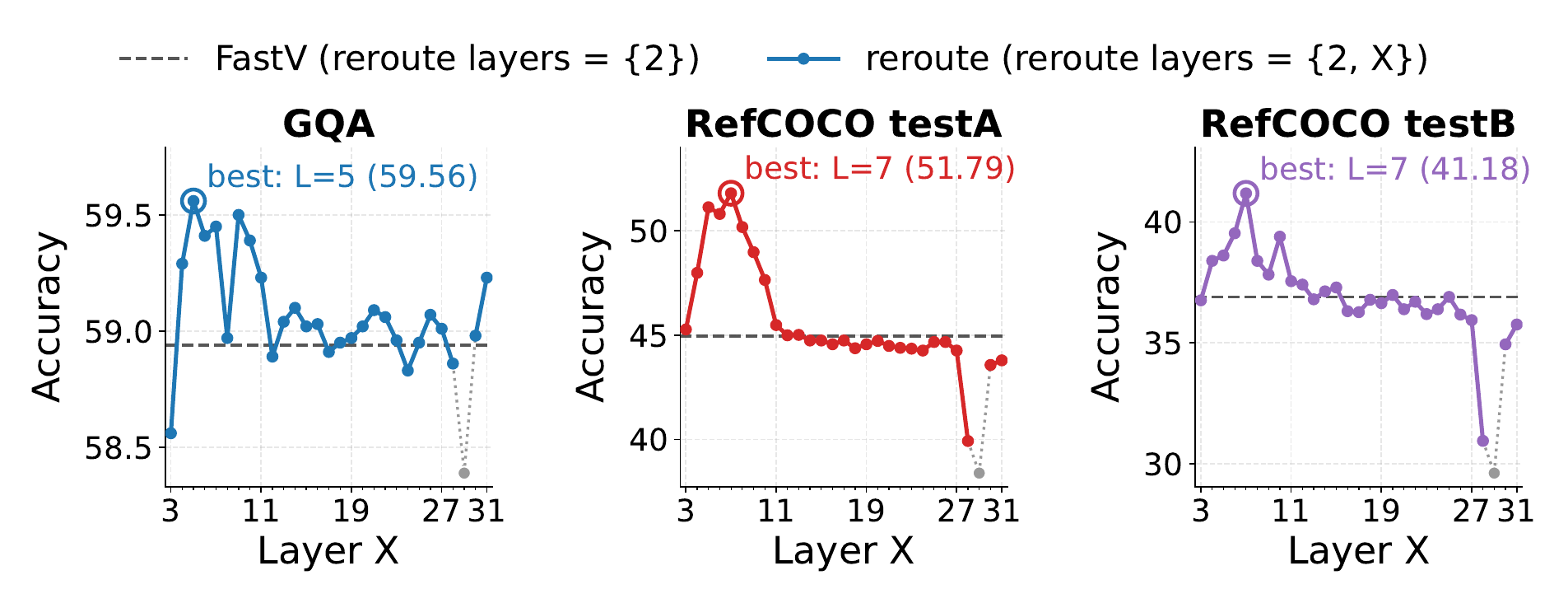}
    \caption{Location sweep.}
    \label{fig:ablation-loc}
  \end{subfigure}
\caption{
\textbf{Ablation of reroute scheduling.}
Under a fixed retention ratio $r_i=0.5$, we vary (a) the number of rerouted layers $N$ and (b) the second reroute location $X$ with the first reroute layer fixed. 
Reroute benefits from repeated re-evaluation, but the best schedule depends on where re-entry is allowed.
}
  \label{fig:ablation-schedule}
  \vspace{-2mm}
\end{figure*}

\vspace{-2mm}
\paragraph{Qwen-series visual grounding results.}
\label{sec:qwen-grounding-results}
We next evaluate whether Reroute transfers beyond LLaVA-1.5. 
Tables~\ref{tab:qwen_refcoco_combined} report RefCOCO-series results on Qwen2.5-VL-7B and Qwen3.5-9B-Hybrid, grouped by average visual-token reduction and gated-attention-token reduction, respectively.

On Qwen2.5-VL, Reroute improves matched pruning baselines across all budgets. 
At 66.7\% reduction, FastV+Reroute improves the average ratio from 53.7\% to 60.3\%, and PDrop+Reroute improves it from 62.9\% to 65.1\%. 
The gains become larger under tighter budgets: PDrop+Reroute improves from 29.3\% to 37.4\% at 77.8\% reduction and from 11.1\% to 18.6\% at 88.9\% reduction.

Qwen3.5-9B-Hybrid requires separate interpretation because it interleaves Gated DeltaNet linear-attention blocks with softmax-attention blocks, and routing is applied only to the latter. Since Gated DeltaNet follows a recurrent state-update view~\cite{yang2025gated}, preserving sequence layout may be important for the remaining hybrid blocks. Reroute is compatible with this setting because deferred tokens bypass routed softmax-attention computation while remaining in the sequence state.

Under this hybrid setting, Reroute yields larger gains. 
At 66.7\% reduction, FastV+Reroute improves the average ratio from 31.1\% to 77.5\%, and PDrop+Reroute from 40.0\% to 77.7\%. 
For PDrop, Reroute further improves the ratio from 25.8\% to 59.7\% at 77.8\% reduction and from 19.3\% to 37.4\% at 88.9\% reduction.

These results reinforce the LLaVA-1.5 finding: Reroute is most beneficial when irreversible pruning prevents later recovery of visual evidence. 
The Qwen3.5 results further suggest that preserving sequence layout may be important for hybrid architectures, although detailed analysis of linear-attention state dynamics is left for future work.

\vspace{-2mm}
\paragraph{Ablation on reroute scheduling.}
\label{sec:ablation-reroute-scheduling}
Figure~\ref{fig:ablation-schedule} ablates the reroute schedule at fixed $r_i=0.5$, varying the number of reroute layers and the placement of an added layer. The single-layer case is exactly FastV; additional reroute layers therefore measure the effect of extending FastV to multi-stage recoverable routing.

The frequency sweep shows that adding reroute stages generally improves over the single-layer FastV reference. On GQA and RefCOCO testA, performance increases with more rerouted layers and peaks under the densest schedule. RefCOCO testB improves up to a moderate frequency, peaking around $N=8$, then saturates. This suggests that repeated re-evaluation helps until deferred tokens have sufficient re-entry opportunities; beyond that, additional routing decisions provide limited benefit and may introduce unnecessary perturbation.

The location sweep shows that Reroute is also sensitive to where the next routing stage is inserted. With the first routing layer fixed, placing the second reroute layer in the early-to-middle decoder performs best: GQA peaks around layer~5, while RefCOCO testA and testB peak around layer~7. Later insertion provides less benefit and can approach the FastV reference. These trends support our stage-wise design and show that the routing schedule should not be treated as arbitrary.

\vspace{-2mm}
\paragraph{Efficiency metrics for Reroute.}
\label{sec:efficiency-metrics}

Table~\ref{tab:efficiency_comparison} reports LLaVA-1.5 efficiency under different average visual-token budgets. 
We report TFLOPs and KV-cache memory from runtime profiler as primary metrics, with ratios normalized by the full-token vanilla baseline.

Reroute largely preserves the efficiency profile of visual-token reduction. 
At the representative 64-token budget, FastV+Reroute reduces main TFLOPs/KV cache by 80\%/86\%, while PDrop+Reroute reduces them by 77\%/83\%, with other budgets remaining close to their matched pruning baselines. 
The small TFLOPs overhead mainly comes from additional reroute stages relative to one-shot FastV, or from the budget-matched schedule relative to PDrop. 
Thus, recoverable routing retains the main compute and cache benefits of pruning; practical wall-clock speedup further depends on optimized gather/scatter and KV-cache implementation.
\vspace{-2mm}
\section{Conclusion}
\label{sec:conclusion}
\vspace{-2mm}

We revisited visual-token reduction from the perspective of the post-scoring action. Instead of permanently removing low-ranked vision tokens, Reroute defers them: selected tokens receive full Attn+FFN computation, while deferred tokens bypass the current stage but remain eligible for later re-entry. Conventional pruning becomes the special case where deferred tokens never return, yielding a training-free Mixture-of-Depth-style formulation for VLM visual tokens.
Empirically, Reroute improves FastV/PDrop-style pruning on LLaVA-1.5 grounding, preserves general VQA, transfers to Qwen-family backbones, and retains the theoretical TFLOPs and KV-cache benefits of token reduction. Gains are largest under aggressive reduction and grounding-oriented evaluation, where irreversible removal is most likely to discard useful evidence. The takeaway is simple: visual-token reduction should preserve the option to recover deferred evidence.



\vspace{-2mm}
\textbf{Limitations.}
Reroute inherits the limitations of its attention-score router: poor ranking or overly aggressive budgets can leave useful evidence inactive for multiple stages. Efficiency gains are primarily theoretical, since practical latency depends on optimized gather/scatter kernels and KV-cache management. Reroute targets decoder-side reduction and is complementary to encoder-side compression, token merging, and KV-cache eviction, whose combinations we do not exhaustively evaluate. Finally, in hybrid architectures such as Qwen3.5, routing is applied only to softmax-attention blocks; interaction with linear-attention state dynamics is left for future work.

\bibliographystyle{plainnat}
\bibliography{example_paper}

\begin{thebibliography}{88}
\providecommand{\natexlab}[1]{#1}
\providecommand{\url}[1]{\texttt{#1}}
\expandafter\ifx\csname urlstyle\endcsname\relax
  \providecommand{\doi}[1]{doi: #1}\else
  \providecommand{\doi}{doi: \begingroup \urlstyle{rm}\Url}\fi

\bibitem[Alvar et~al.(2025)Alvar, Singh, Akbari, and Zhang]{alvar2025divprune}
Saeed~Ranjbar Alvar, Gursimran Singh, Mohammad Akbari, and Yong Zhang.
\newblock Divprune: Diversity-based visual token pruning for large multimodal models.
\newblock In \emph{Proceedings of the Computer Vision and Pattern Recognition Conference}, pages 9392--9401, 2025.

\bibitem[Arif et~al.(2025)Arif, Yoon, Nikolopoulos, Vandierendonck, John, and Ji]{arif2025hired}
Kazi Hasan~Ibn Arif, JinYi Yoon, Dimitrios~S Nikolopoulos, Hans Vandierendonck, Deepu John, and Bo~Ji.
\newblock Hired: Attention-guided token dropping for efficient inference of high-resolution vision-language models.
\newblock In \emph{Proceedings of the AAAI Conference on Artificial Intelligence}, volume~39, pages 1773--1781, 2025.

\bibitem[Bae et~al.(2025)Bae, Kim, Bayat, Kim, Ha, Schuster, Fisch, Harutyunyan, Ji, Courville, et~al.]{bae2025mixture}
Sangmin Bae, Yujin Kim, Reza Bayat, Sungnyun Kim, Jiyoun Ha, Tal Schuster, Adam Fisch, Hrayr Harutyunyan, Ziwei Ji, Aaron Courville, et~al.
\newblock Mixture-of-recursions: Learning dynamic recursive depths for adaptive token-level computation.
\newblock \emph{arXiv preprint arXiv:2507.10524}, 2025.

\bibitem[Bagrov et~al.(2025)Bagrov, Khvedchenia, Tymchenko, Aharon, Kadoch, Keren, Masad, Geifman, Zilberstein, Rintamaki, Le, and Tao]{bagrov2025efficientvideosamplingpruning}
Natan Bagrov, Eugene Khvedchenia, Borys Tymchenko, Shay Aharon, Lior Kadoch, Tomer Keren, Ofri Masad, Yonatan Geifman, Ran Zilberstein, Tuomas Rintamaki, Matthieu Le, and Andrew Tao.
\newblock Efficient video sampling: Pruning temporally redundant tokens for faster vlm inference, 2025.
\newblock URL \url{https://arxiv.org/abs/2510.14624}.

\bibitem[Bai et~al.(2023)Bai, Bai, Yang, Wang, Tan, Wang, Lin, Zhou, and Zhou]{bai2023qwen}
Jinze Bai, Shuai Bai, Shusheng Yang, Shijie Wang, Sinan Tan, Peng Wang, Junyang Lin, Chang Zhou, and Jingren Zhou.
\newblock Qwen-vl: A versatile vision-language model for understanding, localization.
\newblock \emph{Text Reading, and Beyond}, 2\penalty0 (1):\penalty0 1, 2023.

\bibitem[Bai et~al.(2025)Bai, Chen, Liu, Wang, Ge, Song, Dang, Wang, Wang, Tang, Zhong, Zhu, Yang, Li, Wan, Wang, Ding, Fu, Xu, Ye, Zhang, Xie, Cheng, Zhang, Yang, Xu, and Lin]{bai2025qwen25vltechnicalreport}
Shuai Bai, Keqin Chen, Xuejing Liu, Jialin Wang, Wenbin Ge, Sibo Song, Kai Dang, Peng Wang, Shijie Wang, Jun Tang, Humen Zhong, Yuanzhi Zhu, Mingkun Yang, Zhaohai Li, Jianqiang Wan, Pengfei Wang, Wei Ding, Zheren Fu, Yiheng Xu, Jiabo Ye, Xi~Zhang, Tianbao Xie, Zesen Cheng, Hang Zhang, Zhibo Yang, Haiyang Xu, and Junyang Lin.
\newblock Qwen2.5-vl technical report, 2025.
\newblock URL \url{https://arxiv.org/abs/2502.13923}.

\bibitem[Bolya et~al.(2022)Bolya, Fu, Dai, Zhang, Feichtenhofer, and Hoffman]{bolya2022token}
Daniel Bolya, Cheng-Yang Fu, Xiaoliang Dai, Peizhao Zhang, Christoph Feichtenhofer, and Judy Hoffman.
\newblock Token merging: Your vit but faster.
\newblock \emph{arXiv preprint arXiv:2210.09461}, 2022.

\bibitem[Cai et~al.(2024{\natexlab{a}})Cai, Yang, Gao, and Lee]{cai2024matryoshka}
Mu~Cai, Jianwei Yang, Jianfeng Gao, and Yong~Jae Lee.
\newblock Matryoshka multimodal models.
\newblock \emph{arXiv preprint arXiv:2405.17430}, 2024{\natexlab{a}}.

\bibitem[Cai et~al.(2024{\natexlab{b}})Cai, Zhang, Gao, Liu, Li, Liu, Lu, Xiong, Dong, Hu, et~al.]{cai2024pyramidkv}
Zefan Cai, Yichi Zhang, Bofei Gao, Yuliang Liu, Yucheng Li, Tianyu Liu, Keming Lu, Wayne Xiong, Yue Dong, Junjie Hu, et~al.
\newblock Pyramidkv: Dynamic kv cache compression based on pyramidal information funneling.
\newblock \emph{arXiv preprint arXiv:2406.02069}, 2024{\natexlab{b}}.

\bibitem[Cha et~al.(2024)Cha, Kang, Mun, and Roh]{cha2024honeybee}
Junbum Cha, Wooyoung Kang, Jonghwan Mun, and Byungseok Roh.
\newblock Honeybee: Locality-enhanced projector for multimodal llm.
\newblock In \emph{Proceedings of the IEEE/CVF Conference on Computer Vision and Pattern Recognition}, pages 13817--13827, 2024.

\bibitem[Chen et~al.(2024{\natexlab{a}})Chen, Ye, He, Wang, Khashabi, and Yuille]{chen2024efficient}
Jieneng Chen, Luoxin Ye, Ju~He, Zhao-Yang Wang, Daniel Khashabi, and Alan Yuille.
\newblock Efficient large multi-modal models via visual context compression.
\newblock \emph{Advances in neural information processing systems}, 37:\penalty0 73986--74007, 2024{\natexlab{a}}.

\bibitem[Chen et~al.(2024{\natexlab{b}})Chen, Zhao, Liu, Bai, Lin, Zhou, and Chang]{chen2024image}
Liang Chen, Haozhe Zhao, Tianyu Liu, Shuai Bai, Junyang Lin, Chang Zhou, and Baobao Chang.
\newblock An image is worth 1/2 tokens after layer 2: Plug-and-play inference acceleration for large vision-language models.
\newblock In \emph{European Conference on Computer Vision}, pages 19--35. Springer, 2024{\natexlab{b}}.

\bibitem[Chen et~al.(2025)Chen, Xu, Zhang, Liu, Liu, and Liu]{chen2025recoverable}
Yi~Chen, Jian Xu, Xu-Yao Zhang, Wen-Zhuo Liu, Yang-Yang Liu, and Cheng-Lin Liu.
\newblock Recoverable compression: A multimodal vision token recovery mechanism guided by text information.
\newblock In \emph{Proceedings of the AAAI Conference on Artificial Intelligence}, volume~39, pages 2293--2301, 2025.

\bibitem[Chen et~al.(2026)Chen, Shan, Ye, and Chen]{chen2026evoprune}
Yuhao Chen, Bin Shan, Xin Ye, and Cheng Chen.
\newblock Evoprune: Early-stage visual token pruning for efficient mllms.
\newblock \emph{arXiv preprint arXiv:2603.03681}, 2026.

\bibitem[Chien et~al.(2025)Chien, Lin, Tsai, Lai, Chen, and Sun]{chien2025groundingawaretokenpruningrecovering}
Tzu-Chun Chien, Chieh-Kai Lin, Shiang-Feng Tsai, Ruei-Chi Lai, Hung-Jen Chen, and Min Sun.
\newblock Grounding-aware token pruning: Recovering from drastic performance drops in visual grounding caused by pruning, 2025.
\newblock URL \url{https://arxiv.org/abs/2506.21873}.

\bibitem[Dao et~al.(2022)Dao, Fu, Ermon, Rudra, and R{\'e}]{dao2022flashattention}
Tri Dao, Dan Fu, Stefano Ermon, Atri Rudra, and Christopher R{\'e}.
\newblock Flashattention: Fast and memory-efficient exact attention with io-awareness.
\newblock \emph{Advances in neural information processing systems}, 35:\penalty0 16344--16359, 2022.

\bibitem[Endo et~al.(2025)Endo, Wang, and Yeung-Levy]{endo2025feather}
Mark Endo, Xiaohan Wang, and Serena Yeung-Levy.
\newblock Feather the throttle: Revisiting visual token pruning for vision-language model acceleration.
\newblock In \emph{Proceedings of the IEEE/CVF International Conference on Computer Vision}, pages 22826--22835, 2025.

\bibitem[Fu et~al.(2023)Fu, Chen, Shen, Qin, Zhang, Lin, Yang, Zheng, Li, Sun, et~al.]{fu2023mme}
Chaoyou Fu, Peixian Chen, Yunhang Shen, Yulei Qin, Mengdan Zhang, Xu~Lin, Jinrui Yang, Xiawu Zheng, Ke~Li, Xing Sun, et~al.
\newblock Mme: A comprehensive evaluation benchmark for multimodal large language models.
\newblock \emph{arXiv preprint arXiv:2306.13394}, 2023.

\bibitem[Ge et~al.(2023)Ge, Zhang, Liu, Zhang, Han, and Gao]{ge2023model}
Suyu Ge, Yunan Zhang, Liyuan Liu, Minjia Zhang, Jiawei Han, and Jianfeng Gao.
\newblock Model tells you what to discard: Adaptive kv cache compression for llms.
\newblock \emph{arXiv preprint arXiv:2310.01801}, 2023.

\bibitem[Graves(2016)]{graves2016adaptive}
Alex Graves.
\newblock Adaptive computation time for recurrent neural networks.
\newblock \emph{arXiv preprint arXiv:1603.08983}, 2016.

\bibitem[He et~al.(2024)He, Chen, Liu, Shao, Zhou, Zhang, and Zhuang]{he2024zipvl}
Yefei He, Feng Chen, Jing Liu, Wenqi Shao, Hong Zhou, Kaipeng Zhang, and Bohan Zhuang.
\newblock Zipvl: Efficient large vision-language models with dynamic token sparsification.
\newblock \emph{arXiv preprint arXiv:2410.08584}, 2024.

\bibitem[Hu et~al.(2024{\natexlab{a}})Hu, Gao, Shang, Wan, and Feng]{hu2024illava}
Lianyu Hu, Liqing Gao, Fanhua Shang, Liang Wan, and Wei Feng.
\newblock illava: An image is worth fewer than 1/3 input tokens in large multimodal models.
\newblock \emph{arXiv preprint arXiv:2412.06263}, 2024{\natexlab{a}}.

\bibitem[Hu et~al.(2024{\natexlab{b}})Hu, Dou, Li, Kamath, Peng, and Chang]{hu2024matryoshka}
Wenbo Hu, Zi-Yi Dou, Liunian~H Li, Amita Kamath, Nanyun Peng, and Kai-Wei Chang.
\newblock Matryoshka query transformer for large vision-language models.
\newblock \emph{Advances in Neural Information Processing Systems}, 37:\penalty0 50168--50188, 2024{\natexlab{b}}.

\bibitem[Huang et~al.(2017)Huang, Chen, Li, Wu, Van Der~Maaten, and Weinberger]{huang2017multi}
Gao Huang, Danlu Chen, Tianhong Li, Felix Wu, Laurens Van Der~Maaten, and Kilian~Q Weinberger.
\newblock Multi-scale dense networks for resource efficient image classification.
\newblock \emph{arXiv preprint arXiv:1703.09844}, 2017.

\bibitem[Huang et~al.(2024)Huang, Zhai, Shen, Cao, Zhao, Xu, Ye, Hu, and Lin]{huang2024dynamic}
Wenxuan Huang, Zijie Zhai, Yunhang Shen, Shaosheng Cao, Fei Zhao, Xiangfeng Xu, Zheyu Ye, Yao Hu, and Shaohui Lin.
\newblock Dynamic-llava: Efficient multimodal large language models via dynamic vision-language context sparsification.
\newblock \emph{arXiv preprint arXiv:2412.00876}, 2024.

\bibitem[Huang et~al.(2026)Huang, Ma, Shao, Guo, Yu, Cui, and Tian]{huang2026n}
Yihong Huang, Fei Ma, Yihua Shao, Jingcai Guo, Zitong Yu, Laizhong Cui, and Qi~Tian.
\newblock N$\backslash$" uwa: Mending the spatial integrity torn by vlm token pruning.
\newblock \emph{arXiv preprint arXiv:2602.02951}, 2026.

\bibitem[Hudson and Manning(2019)]{hudson2019gqa}
Drew~A Hudson and Christopher~D Manning.
\newblock Gqa: A new dataset for real-world visual reasoning and compositional question answering.
\newblock In \emph{Proceedings of the IEEE/CVF conference on computer vision and pattern recognition}, pages 6700--6709, 2019.

\bibitem[Jiang et~al.(2024)Jiang, Huang, Liu, Zeng, Li, Cheng, and Xu]{jiang2024fopru}
Lei Jiang, Weizhe Huang, Tongxuan Liu, Yuting Zeng, Jing Li, Lechao Cheng, and Xiaohua Xu.
\newblock Fopru: Focal pruning for efficient large vision-language models.
\newblock \emph{arXiv preprint arXiv:2411.14164}, 2024.

\bibitem[Kang et~al.(2025)Kang, Kim, Kim, and Hwang]{kang2025see}
Seil Kang, Jinyeong Kim, Junhyeok Kim, and Seong~Jae Hwang.
\newblock See what you are told: Visual attention sink in large multimodal models.
\newblock \emph{arXiv preprint arXiv:2503.03321}, 2025.

\bibitem[Kazemzadeh et~al.(2014)Kazemzadeh, Ordonez, Matten, and Berg]{kazemzadeh2014referitgame}
Sahar Kazemzadeh, Vicente Ordonez, Mark Matten, and Tamara Berg.
\newblock Referitgame: Referring to objects in photographs of natural scenes.
\newblock In \emph{Proceedings of the 2014 conference on empirical methods in natural language processing (EMNLP)}, pages 787--798, 2014.

\bibitem[Kwon et~al.(2023)Kwon, Li, Zhuang, Sheng, Zheng, Yu, Gonzalez, Zhang, and Stoica]{kwon2023efficient}
Woosuk Kwon, Zhuohan Li, Siyuan Zhuang, Ying Sheng, Lianmin Zheng, Cody~Hao Yu, Joseph Gonzalez, Hao Zhang, and Ion Stoica.
\newblock Efficient memory management for large language model serving with pagedattention.
\newblock In \emph{Proceedings of the 29th symposium on operating systems principles}, pages 611--626, 2023.

\bibitem[Lawson and Aitchison(2025)]{lawson2025learning}
Tim Lawson and Laurence Aitchison.
\newblock Learning to skip the middle layers of transformers.
\newblock \emph{arXiv preprint arXiv:2506.21103}, 2025.

\bibitem[Li et~al.(2023{\natexlab{a}})Li, Wang, Wang, Ge, Ge, and Shan]{li2023seed}
Bohao Li, Rui Wang, Guangzhi Wang, Yuying Ge, Yixiao Ge, and Ying Shan.
\newblock Seed-bench: Benchmarking multimodal llms with generative comprehension.
\newblock \emph{arXiv preprint arXiv:2307.16125}, 2023{\natexlab{a}}.

\bibitem[Li et~al.(2024{\natexlab{a}})Li, Wang, Zhu, Kuo, Xu, Chen, Jain, Shi, and Wen]{li2024cumo}
Jiachen Li, Xinyao Wang, Sijie Zhu, Chia-Wen Kuo, Lu~Xu, Fan Chen, Jitesh Jain, Humphrey Shi, and Longyin Wen.
\newblock Cumo: Scaling multimodal llm with co-upcycled mixture-of-experts.
\newblock \emph{Advances in Neural Information Processing Systems}, 37:\penalty0 131224--131246, 2024{\natexlab{a}}.

\bibitem[Li et~al.(2025)Li, Yuan, Liu, Tang, Wang, Qin, Zhu, and Zhang]{li2025tokenpacker}
Wentong Li, Yuqian Yuan, Jian Liu, Dongqi Tang, Song Wang, Jie Qin, Jianke Zhu, and Lei Zhang.
\newblock Tokenpacker: Efficient visual projector for multimodal llm.
\newblock \emph{International Journal of Computer Vision}, 133\penalty0 (10):\penalty0 6794--6812, 2025.

\bibitem[Li et~al.(2023{\natexlab{b}})Li, Du, Zhou, Wang, Zhao, and Wen]{li2023evaluating}
Yifan Li, Yifan Du, Kun Zhou, Jinpeng Wang, Xin Zhao, and Ji-Rong Wen.
\newblock Evaluating object hallucination in large vision-language models.
\newblock In \emph{Proceedings of the 2023 conference on empirical methods in natural language processing}, pages 292--305, 2023{\natexlab{b}}.

\bibitem[Li et~al.(2024{\natexlab{b}})Li, Huang, Yang, Venkitesh, Locatelli, Ye, Cai, Lewis, and Chen]{li2024snapkv}
Yuhong Li, Yingbing Huang, Bowen Yang, Bharat Venkitesh, Acyr Locatelli, Hanchen Ye, Tianle Cai, Patrick Lewis, and Deming Chen.
\newblock Snapkv: Llm knows what you are looking for before generation.
\newblock \emph{Advances in Neural Information Processing Systems}, 37:\penalty0 22947--22970, 2024{\natexlab{b}}.

\bibitem[Liang et~al.(2022)Liang, Ge, Tong, Song, Wang, and Xie]{liang2022not}
Youwei Liang, Chongjian Ge, Zhan Tong, Yibing Song, Jue Wang, and Pengtao Xie.
\newblock Not all patches are what you need: Expediting vision transformers via token reorganizations.
\newblock \emph{arXiv preprint arXiv:2202.07800}, 2022.

\bibitem[Lin et~al.(2026)Lin, Tang, Ye, Huang, Zhang, Pang, Jin, Ning, Luo, and Yuan]{lin2026moe}
Bin Lin, Zhenyu Tang, Yang Ye, Jinfa Huang, Junwu Zhang, Yatian Pang, Peng Jin, Munan Ning, Jiebo Luo, and Li~Yuan.
\newblock Moe-llava: Mixture of experts for large vision-language models.
\newblock \emph{IEEE Transactions on Multimedia}, 2026.

\bibitem[Lin et~al.(2025)Lin, Lin, Lin, and Ji]{lin2025boosting}
Zhihang Lin, Mingbao Lin, Luxi Lin, and Rongrong Ji.
\newblock Boosting multimodal large language models with visual tokens withdrawal for rapid inference.
\newblock In \emph{Proceedings of the AAAI Conference on Artificial Intelligence}, volume~39, pages 5334--5342, 2025.

\bibitem[Liu et~al.(2023{\natexlab{a}})Liu, Li, Wu, and Lee]{liu2023visual}
Haotian Liu, Chunyuan Li, Qingyang Wu, and Yong~Jae Lee.
\newblock Visual instruction tuning.
\newblock \emph{Advances in neural information processing systems}, 36:\penalty0 34892--34916, 2023{\natexlab{a}}.

\bibitem[Liu et~al.(2024{\natexlab{a}})Liu, Li, Li, and Lee]{liu2024improved}
Haotian Liu, Chunyuan Li, Yuheng Li, and Yong~Jae Lee.
\newblock Improved baselines with visual instruction tuning.
\newblock In \emph{Proceedings of the IEEE/CVF conference on computer vision and pattern recognition}, pages 26296--26306, 2024{\natexlab{a}}.

\bibitem[Liu et~al.(2026)Liu, Zhu, and Du]{liu2026hiprune}
Jizhihui Liu, Guangdao Zhu, and Feiyi Du.
\newblock Hiprune: Training-free visual token pruning via hierarchical attention in vision-language models (student abstract).
\newblock In \emph{Proceedings of the AAAI Conference on Artificial Intelligence}, volume~40, pages 41275--41277, 2026.

\bibitem[Liu et~al.(2024{\natexlab{b}})Liu, Shi, Hong, Hu, Yin, and Zhang]{liu2024multi}
Ting Liu, Liangtao Shi, Richang Hong, Yue Hu, Quanjun Yin, and Linfeng Zhang.
\newblock Multi-stage vision token dropping: Towards efficient multimodal large language model.
\newblock \emph{arXiv preprint arXiv:2411.10803}, 2024{\natexlab{b}}.

\bibitem[Liu et~al.(2020)Liu, Zhou, Wang, Zhao, Deng, and Ju]{liu2020fastbert}
Weijie Liu, Peng Zhou, Zhiruo Wang, Zhe Zhao, Haotang Deng, and Qi~Ju.
\newblock Fastbert: a self-distilling bert with adaptive inference time.
\newblock In \emph{Proceedings of the 58th annual meeting of the association for computational linguistics}, pages 6035--6044, 2020.

\bibitem[Liu et~al.(2024{\natexlab{c}})Liu, Duan, Zhang, Li, Zhang, Zhao, Yuan, Wang, He, Liu, et~al.]{liu2024mmbench}
Yuan Liu, Haodong Duan, Yuanhan Zhang, Bo~Li, Songyang Zhang, Wangbo Zhao, Yike Yuan, Jiaqi Wang, Conghui He, Ziwei Liu, et~al.
\newblock Mmbench: Is your multi-modal model an all-around player?
\newblock In \emph{European conference on computer vision}, pages 216--233. Springer, 2024{\natexlab{c}}.

\bibitem[Liu et~al.(2023{\natexlab{b}})Liu, Desai, Liao, Wang, Xie, Xu, Kyrillidis, and Shrivastava]{liu2023scissorhands}
Zichang Liu, Aditya Desai, Fangshuo Liao, Weitao Wang, Victor Xie, Zhaozhuo Xu, Anastasios Kyrillidis, and Anshumali Shrivastava.
\newblock Scissorhands: Exploiting the persistence of importance hypothesis for llm kv cache compression at test time.
\newblock \emph{Advances in Neural Information Processing Systems}, 36:\penalty0 52342--52364, 2023{\natexlab{b}}.

\bibitem[Liu et~al.(2024{\natexlab{d}})Liu, Liu, Wang, Dong, Chen, Rao, Krishna, and Lu]{liu2024efficient}
Zuyan Liu, Benlin Liu, Jiahui Wang, Yuhao Dong, Guangyi Chen, Yongming Rao, Ranjay Krishna, and Jiwen Lu.
\newblock Efficient inference of vision instruction-following models with elastic cache.
\newblock In \emph{European Conference on Computer Vision}, pages 54--69. Springer, 2024{\natexlab{d}}.

\bibitem[Lu et~al.(2022)Lu, Mishra, Xia, Qiu, Chang, Zhu, Tafjord, Clark, and Kalyan]{lu2022learn}
Pan Lu, Swaroop Mishra, Tanglin Xia, Liang Qiu, Kai-Wei Chang, Song-Chun Zhu, Oyvind Tafjord, Peter Clark, and Ashwin Kalyan.
\newblock Learn to explain: Multimodal reasoning via thought chains for science question answering.
\newblock \emph{Advances in neural information processing systems}, 35:\penalty0 2507--2521, 2022.

\bibitem[Luo et~al.(2024)Luo, Luo, Ji, Zhou, Sun, Shen, and Ji]{luo2024gamma}
Yaxin Luo, Gen Luo, Jiayi Ji, Yiyi Zhou, Xiaoshuai Sun, Zhiqiang Shen, and Rongrong Ji.
\newblock $\gamma-$mod: Exploring mixture-of-depth adaptation for multimodal large language models.
\newblock \emph{arXiv preprint arXiv:2410.13859}, 2024.

\bibitem[Mahmud et~al.(2024)Mahmud, Yaman, Liu, and Marculescu]{mahmud2024papr}
Tanvir Mahmud, Burhaneddin Yaman, Chun-Hao Liu, and Diana Marculescu.
\newblock Papr: Training-free one-step patch pruning with lightweight convnets for faster inference.
\newblock In \emph{European Conference on Computer Vision}, pages 110--128. Springer, 2024.

\bibitem[Meng et~al.(2024)Meng, Yang, Tian, Dai, Wu, Gao, and Jiang]{meng2024deepstack}
Lingchen Meng, Jianwei Yang, Rui Tian, Xiyang Dai, Zuxuan Wu, Jianfeng Gao, and Yu-Gang Jiang.
\newblock Deepstack: Deeply stacking visual tokens is surprisingly simple and effective for lmms.
\newblock \emph{Advances in Neural Information Processing Systems}, 37:\penalty0 23464--23487, 2024.

\bibitem[Olsson et~al.(2022)Olsson, Elhage, Nanda, Joseph, DasSarma, Henighan, Mann, Askell, Bai, Chen, et~al.]{olsson2022context}
Catherine Olsson, Nelson Elhage, Neel Nanda, Nicholas Joseph, Nova DasSarma, Tom Henighan, Ben Mann, Amanda Askell, Yuntao Bai, Anna Chen, et~al.
\newblock In-context learning and induction heads.
\newblock \emph{arXiv preprint arXiv:2209.11895}, 2022.

\bibitem[{Qwen Team}(2026)]{qwen3.5}
{Qwen Team}.
\newblock {Qwen3.5}: Towards native multimodal agents, February 2026.
\newblock URL \url{https://qwen.ai/blog?id=qwen3.5}.

\bibitem[Rao et~al.(2021)Rao, Zhao, Liu, Lu, Zhou, and Hsieh]{rao2021dynamicvit}
Yongming Rao, Wenliang Zhao, Benlin Liu, Jiwen Lu, Jie Zhou, and Cho-Jui Hsieh.
\newblock Dynamicvit: Efficient vision transformers with dynamic token sparsification.
\newblock \emph{Advances in neural information processing systems}, 34:\penalty0 13937--13949, 2021.

\bibitem[Raposo et~al.(2024)Raposo, Ritter, Richards, Lillicrap, Humphreys, and Santoro]{raposo2024mixture}
David Raposo, Sam Ritter, Blake Richards, Timothy Lillicrap, Peter~Conway Humphreys, and Adam Santoro.
\newblock Mixture-of-depths: Dynamically allocating compute in transformer-based language models.
\newblock \emph{arXiv preprint arXiv:2404.02258}, 2024.

\bibitem[Schuster et~al.(2022)Schuster, Fisch, Gupta, Dehghani, Bahri, Tran, Tay, and Metzler]{schuster2022confident}
Tal Schuster, Adam Fisch, Jai Gupta, Mostafa Dehghani, Dara Bahri, Vinh Tran, Yi~Tay, and Donald Metzler.
\newblock Confident adaptive language modeling.
\newblock \emph{Advances in Neural Information Processing Systems}, 35:\penalty0 17456--17472, 2022.

\bibitem[Shang et~al.(2025)Shang, Cai, Xu, Lee, and Yan]{shang2025llava}
Yuzhang Shang, Mu~Cai, Bingxin Xu, Yong~Jae Lee, and Yan Yan.
\newblock Llava-prumerge: Adaptive token reduction for efficient large multimodal models.
\newblock In \emph{Proceedings of the IEEE/CVF International Conference on Computer Vision}, pages 22857--22867, 2025.

\bibitem[Singh et~al.(2019)Singh, Natarajan, Shah, Jiang, Chen, Batra, Parikh, and Rohrbach]{singh2019towards}
Amanpreet Singh, Vivek Natarajan, Meet Shah, Yu~Jiang, Xinlei Chen, Dhruv Batra, Devi Parikh, and Marcus Rohrbach.
\newblock Towards vqa models that can read.
\newblock In \emph{Proceedings of the IEEE/CVF conference on computer vision and pattern recognition}, pages 8317--8326, 2019.

\bibitem[Sun et~al.(2024)Sun, Chen, Kolter, and Liu]{sun2024massive}
Mingjie Sun, Xinlei Chen, J~Zico Kolter, and Zhuang Liu.
\newblock Massive activations in large language models.
\newblock \emph{arXiv preprint arXiv:2402.17762}, 2024.

\bibitem[Teerapittayanon et~al.(2016)Teerapittayanon, McDanel, and Kung]{teerapittayanon2016branchynet}
Surat Teerapittayanon, Bradley McDanel, and Hsiang-Tsung Kung.
\newblock Branchynet: Fast inference via early exiting from deep neural networks.
\newblock In \emph{2016 23rd international conference on pattern recognition (ICPR)}, pages 2464--2469. IEEE, 2016.

\bibitem[Wan et~al.(2024)Wan, Wu, Liu, Huang, Zhu, Jin, Wang, and Yuan]{wan2024look}
Zhongwei Wan, Ziang Wu, Che Liu, Jinfa Huang, Zhihong Zhu, Peng Jin, Longyue Wang, and Li~Yuan.
\newblock Look-m: Look-once optimization in kv cache for efficient multimodal long-context inference.
\newblock In \emph{Findings of the Association for Computational Linguistics: EMNLP 2024}, pages 4065--4078, 2024.

\bibitem[Wang et~al.(2025)Wang, Yu, Spadaro, Ju, Qu{\'e}tu, Xiao, and Tartaglione]{wang2025folder}
Haicheng Wang, Zhemeng Yu, Gabriele Spadaro, Chen Ju, Victor Qu{\'e}tu, Shuai Xiao, and Enzo Tartaglione.
\newblock Folder: Accelerating multi-modal large language models with enhanced performance.
\newblock In \emph{Proceedings of the IEEE/CVF International Conference on Computer Vision}, pages 23614--23625, 2025.

\bibitem[Wang et~al.(2018)Wang, Yu, Dou, Darrell, and Gonzalez]{wang2018skipnet}
Xin Wang, Fisher Yu, Zi-Yi Dou, Trevor Darrell, and Joseph~E Gonzalez.
\newblock Skipnet: Learning dynamic routing in convolutional networks.
\newblock In \emph{Proceedings of the European conference on computer vision (ECCV)}, pages 409--424, 2018.

\bibitem[Wu et~al.(2024{\natexlab{a}})Wu, Lin, Zhou, Ye, Zen, Sun, and Ji]{wu2024accelerating}
Qiong Wu, Wenhao Lin, Yiyi Zhou, Weihao Ye, Zhanpeng Zen, Xiaoshuai Sun, and Rongrong Ji.
\newblock Accelerating multimodal large language models via dynamic visual-token exit and the empirical findings.
\newblock \emph{arXiv preprint arXiv:2411.19628}, 2024{\natexlab{a}}.

\bibitem[Wu et~al.(2025)Wu, Ke, Zhou, Sun, and Ji]{wu2025routing}
Qiong Wu, Zhaoxi Ke, Yiyi Zhou, Xiaoshuai Sun, and Rongrong Ji.
\newblock Routing experts: Learning to route dynamic experts in existing multi-modal large language models.
\newblock In \emph{The Thirteenth International Conference on Learning Representations}, 2025.

\bibitem[Wu et~al.(2024{\natexlab{b}})Wu, Wang, Xiao, Peng, and Fu]{wu2024retrieval}
Wenhao Wu, Yizhong Wang, Guangxuan Xiao, Hao Peng, and Yao Fu.
\newblock Retrieval head mechanistically explains long-context factuality.
\newblock \emph{arXiv preprint arXiv:2404.15574}, 2024{\natexlab{b}}.

\bibitem[Wu et~al.(2018)Wu, Nagarajan, Kumar, Rennie, Davis, Grauman, and Feris]{wu2018blockdrop}
Zuxuan Wu, Tushar Nagarajan, Abhishek Kumar, Steven Rennie, Larry~S Davis, Kristen Grauman, and Rogerio Feris.
\newblock Blockdrop: Dynamic inference paths in residual networks.
\newblock In \emph{Proceedings of the IEEE conference on computer vision and pattern recognition}, pages 8817--8826, 2018.

\bibitem[Xiao et~al.(2023)Xiao, Tian, Chen, Han, and Lewis]{xiao2023efficient}
Guangxuan Xiao, Yuandong Tian, Beidi Chen, Song Han, and Mike Lewis.
\newblock Efficient streaming language models with attention sinks.
\newblock \emph{arXiv preprint arXiv:2309.17453}, 2023.

\bibitem[Xin et~al.(2020)Xin, Tang, Lee, Yu, and Lin]{xin2020deebert}
Ji~Xin, Raphael Tang, Jaejun Lee, Yaoliang Yu, and Jimmy Lin.
\newblock Deebert: Dynamic early exiting for accelerating bert inference.
\newblock In \emph{Proceedings of the 58th annual meeting of the association for computational linguistics}, pages 2246--2251, 2020.

\bibitem[Xing et~al.(2024)Xing, Huang, Dong, Lu, Zhang, Zang, Cao, He, Wang, Wu, et~al.]{xing2024pyramiddrop}
Long Xing, Qidong Huang, Xiaoyi Dong, Jiajie Lu, Pan Zhang, Yuhang Zang, Yuhang Cao, Conghui He, Jiaqi Wang, Feng Wu, et~al.
\newblock Pyramiddrop: Accelerating your large vision-language models via pyramid visual redundancy reduction.
\newblock \emph{arXiv preprint arXiv:2410.17247}, 2024.

\bibitem[Yang et~al.(2025{\natexlab{a}})Yang, Sui, Xiao, Huang, Gong, Li, Yan, Bai, Sadayappan, Hu, et~al.]{yang2025topv}
Cheng Yang, Yang Sui, Jinqi Xiao, Lingyi Huang, Yu~Gong, Chendi Li, Jinghua Yan, Yu~Bai, Ponnuswamy Sadayappan, Xia Hu, et~al.
\newblock Topv: Compatible token pruning with inference time optimization for fast and low-memory multimodal vision language model.
\newblock In \emph{Proceedings of the Computer Vision and Pattern Recognition Conference}, pages 19803--19813, 2025{\natexlab{a}}.

\bibitem[Yang et~al.(2025{\natexlab{b}})Yang, Chen, Tian, Wang, Li, Yu, and Jia]{yang2025visionzip}
Senqiao Yang, Yukang Chen, Zhuotao Tian, Chengyao Wang, Jingyao Li, Bei Yu, and Jiaya Jia.
\newblock Visionzip: Longer is better but not necessary in vision language models.
\newblock In \emph{Proceedings of the IEEE/CVF Conference on Computer Vision and Pattern Recognition}, pages 19792--19802, 2025{\natexlab{b}}.

\bibitem[Yang et~al.(2025{\natexlab{c}})Yang, Kautz, and Hatamizadeh]{yang2025gated}
Songlin Yang, Jan Kautz, and Ali Hatamizadeh.
\newblock Gated delta networks: Improving mamba2 with delta rule.
\newblock In \emph{The Thirteenth International Conference on Learning Representations}, 2025{\natexlab{c}}.
\newblock URL \url{https://openreview.net/forum?id=r8H7xhYPwz}.

\bibitem[Ye et~al.(2025{\natexlab{a}})Ye, Wu, Lin, and Zhou]{ye2025fit}
Weihao Ye, Qiong Wu, Wenhao Lin, and Yiyi Zhou.
\newblock Fit and prune: Fast and training-free visual token pruning for multi-modal large language models.
\newblock In \emph{Proceedings of the AAAI Conference on Artificial Intelligence}, volume~39, pages 22128--22136, 2025{\natexlab{a}}.

\bibitem[Ye et~al.(2025{\natexlab{b}})Ye, Gan, Ge, Zhang, and Tang]{ye2025atp}
Xubing Ye, Yukang Gan, Yixiao Ge, Xiao-Ping Zhang, and Yansong Tang.
\newblock Atp-llava: Adaptive token pruning for large vision language models.
\newblock In \emph{Proceedings of the IEEE/CVF Conference on Computer Vision and Pattern Recognition}, pages 24972--24982, 2025{\natexlab{b}}.

\bibitem[Ye et~al.(2025{\natexlab{c}})Ye, Gan, Huang, Ge, and Tang]{ye2025voco}
Xubing Ye, Yukang Gan, Xiaoke Huang, Yixiao Ge, and Yansong Tang.
\newblock Voco-llama: Towards vision compression with large language models.
\newblock In \emph{Proceedings of the Computer Vision and Pattern Recognition Conference}, pages 29836--29846, 2025{\natexlab{c}}.

\bibitem[Yin et~al.(2025)Yin, Si, and Wang]{yin2025lifting}
Hao Yin, Guangzong Si, and Zilei Wang.
\newblock Lifting the veil on visual information flow in mllms: unlocking pathways to faster inference.
\newblock In \emph{Proceedings of the IEEE/CVF Conference on Computer Vision and Pattern Recognition}, pages 9382--9391, 2025.

\bibitem[Yu et~al.(2016)Yu, Poirson, Yang, Berg, and Berg]{yu2016modeling}
Licheng Yu, Patrick Poirson, Shan Yang, Alexander~C Berg, and Tamara~L Berg.
\newblock Modeling context in referring expressions.
\newblock In \emph{European conference on computer vision}, pages 69--85. Springer, 2016.

\bibitem[Yue et~al.(2024)Yue, Ni, Zhang, Zheng, Liu, Zhang, Stevens, Jiang, Ren, Sun, et~al.]{yue2024mmmu}
Xiang Yue, Yuansheng Ni, Kai Zhang, Tianyu Zheng, Ruoqi Liu, Ge~Zhang, Samuel Stevens, Dongfu Jiang, Weiming Ren, Yuxuan Sun, et~al.
\newblock Mmmu: A massive multi-discipline multimodal understanding and reasoning benchmark for expert agi.
\newblock In \emph{Proceedings of the IEEE/CVF conference on computer vision and pattern recognition}, pages 9556--9567, 2024.

\bibitem[Zhang et~al.(2025{\natexlab{a}})Zhang, Meng, Zhang, Huang, Wu, and Wang]{zhang2025p}
Jun Zhang, Desen Meng, Zhengming Zhang, Zhenpeng Huang, Tao Wu, and Limin Wang.
\newblock p-mod: Building mixture-of-depths mllms via progressive ratio decay.
\newblock In \emph{Proceedings of the IEEE/CVF International Conference on Computer Vision}, pages 3705--3715, 2025{\natexlab{a}}.

\bibitem[Zhang et~al.(2024{\natexlab{a}})Zhang, Cheng, Lu, Zhuo, Wang, Cao, Guo, She, and Zhang]{zhang2024cls}
Qizhe Zhang, Aosong Cheng, Ming Lu, Zhiyong Zhuo, Minqi Wang, Jiajun Cao, Shaobo Guo, Qi~She, and Shanghang Zhang.
\newblock [cls] attention is all you need for training-free visual token pruning: Make vlm inference faster.
\newblock \emph{arXiv e-prints}, pages arXiv--2412, 2024{\natexlab{a}}.

\bibitem[Zhang et~al.(2025{\natexlab{b}})Zhang, Cheng, Lu, Zhang, Zhuo, Cao, Guo, She, and Zhang]{zhang2025beyond}
Qizhe Zhang, Aosong Cheng, Ming Lu, Renrui Zhang, Zhiyong Zhuo, Jiajun Cao, Shaobo Guo, Qi~She, and Shanghang Zhang.
\newblock Beyond text-visual attention: Exploiting visual cues for effective token pruning in vlms.
\newblock In \emph{Proceedings of the IEEE/CVF International Conference on Computer Vision}, pages 20857--20867, 2025{\natexlab{b}}.

\bibitem[Zhang et~al.(2024{\natexlab{b}})Zhang, Fan, Ma, Zheng, Huang, Cheng, Gudovskiy, Okuno, Nakata, Keutzer, et~al.]{zhang2024sparsevlm}
Yuan Zhang, Chun-Kai Fan, Junpeng Ma, Wenzhao Zheng, Tao Huang, Kuan Cheng, Denis Gudovskiy, Tomoyuki Okuno, Yohei Nakata, Kurt Keutzer, et~al.
\newblock Sparsevlm: Visual token sparsification for efficient vision-language model inference.
\newblock \emph{arXiv preprint arXiv:2410.04417}, 2024{\natexlab{b}}.

\bibitem[Zhang et~al.(2023)Zhang, Sheng, Zhou, Chen, Zheng, Cai, Song, Tian, R{\'e}, Barrett, et~al.]{zhang2023h2o}
Zhenyu Zhang, Ying Sheng, Tianyi Zhou, Tianlong Chen, Lianmin Zheng, Ruisi Cai, Zhao Song, Yuandong Tian, Christopher R{\'e}, Clark Barrett, et~al.
\newblock H2o: Heavy-hitter oracle for efficient generative inference of large language models.
\newblock \emph{Advances in Neural Information Processing Systems}, 36:\penalty0 34661--34710, 2023.

\bibitem[Zhao et~al.(2025)Zhao, Zhang, Li, Xing, Yuan, Tang, Fan, Chen, Wang, and Zhang]{zhao2025mca}
Qiyan Zhao, Xiaofeng Zhang, Yiheng Li, Yun Xing, Xiaosong Yuan, Feilong Tang, Sinan Fan, Xuhang Chen, Da-Han Wang, and Xu-Yao Zhang.
\newblock Mca-llava: Manhattan causal attention for reducing hallucination in large vision-language models.
\newblock In \emph{Proceedings of the 33rd ACM International Conference on Multimedia}, pages 3981--3990, 2025.

\bibitem[Zhuang et~al.(2025)Zhuang, Lu, Dai, Hu, Chen, Liu, and Hu]{zhuang2025st3}
Jiedong Zhuang, Lu~Lu, Ming Dai, Rui Hu, Jian Chen, Qiang Liu, and Haoji Hu.
\newblock St3: Accelerating multimodal large language model by spatial-temporal visual token trimming.
\newblock In \emph{Proceedings of the AAAI Conference on Artificial Intelligence}, volume~39, pages 11049--11057, 2025.

\bibitem[Zong et~al.(2024)Zong, Ma, Shen, Song, Shao, Jiang, Li, and Liu]{zong2024mova}
Zhuofan Zong, Bingqi Ma, Dazhong Shen, Guanglu Song, Hao Shao, Dongzhi Jiang, Hongsheng Li, and Yu~Liu.
\newblock Mova: Adapting mixture of vision experts to multimodal context.
\newblock \emph{Advances in Neural Information Processing Systems}, 37:\penalty0 103305--103333, 2024.

\end{thebibliography}


\newpage
\appendix




\appendix

\section*{Appendix Overview}

This appendix supplements the main paper with reproducibility details, routing-schedule analyses, and additional results. Appendix~\ref{app:protocol-config} specifies the datasets, splits, model-specific inference settings, routing configurations, and position-index handling used in our experiments. Appendix~\ref{app:schedule-analysis} analyzes the schedule choices behind the main Reroute configurations, including the early-stage PDrop ablation and the frequency/location sweeps for reroute scheduling. Appendix~\ref{app:supp-results} reports additional VQA preservation results, discusses Qwen3-VL as an architecture-sensitive grounding case study, and provides qualitative grounding and token-selection visualizations. Together, these materials document how the reported results are reproduced and further analyze when recoverable routing is most beneficial.

\section{Experimental Protocol and Configurations}
\label{app:protocol-config}

This section provides the implementation and evaluation details needed to reproduce the main results. We first describe the datasets, evaluation splits, model settings, and grounding post-processing conventions. We then specify the routing configurations, cross-model schedule normalization, and position-index handling used throughout the experiments.

\subsection{Evaluation datasets and splits}
\label{app:datasets-splits}

For visual grounding, we evaluate on the RefCOCO-family grounding suite in \texttt{lmms-eval}. The evaluated splits include RefCOCO (\texttt{refcoco\_val}, \texttt{refcoco\_testA}, \texttt{refcoco\_testB}), RefCOCO+ (\texttt{refcoco+\_val}, \texttt{refcoco+\_testA}, \texttt{refcoco+\_testB}), and RefCOCOg (\texttt{refcocog\_val}, \texttt{refcocog\_test}). These benchmarks evaluate whether a model can localize the object referred to by a natural-language expression. The model prediction is parsed as a bounding box and evaluated using the standard IoU-based grounding protocol implemented by \texttt{lmms-eval}.

To verify that visual-token reduction does not improve grounding at the expense of general multimodal understanding, we also evaluate on a general VQA suite. The core suite includes MME, GQA, POPE, TextVQA, ScienceQA, and MMBench. When available, we additionally report results on AI2D, MMMU-Val, VQAv2-Lite, MMStar, and SEEDBench-Lite. These tasks cover general perception, object recognition, hallucination evaluation, OCR-oriented question answering, scientific reasoning, and broader multimodal reasoning.

\subsection{Model and inference settings}
\label{app:model-inference-settings}

We evaluate Reroute and its pruning baselines on LLaVA-1.5-7B, Qwen2.5-VL-7B, and Qwen3.5-9B. We additionally include Qwen3-VL-8B for architecture-level discussion and supplementary comparisons. Unless otherwise noted, each benchmark uses the default generation configuration provided by \texttt{lmms-eval}. All experiments are conducted on a single NVIDIA RTX 4090 GPU.

For visual grounding, prompt and post-processing conventions follow the corresponding model family. For LLaVA-1.5, we use the default \texttt{lmms-eval} prompt and post-processing pipeline. For Qwen-family models, we follow the official/Nuwa prompt format and the community-suggested grounding post-processing convention. In particular, Qwen2.5-VL bounding boxes are normalized by pixel coordinates, which is already handled by \texttt{lmms-eval}, whereas Qwen3-VL and Qwen3.5 use $[0,1000]$ coordinate normalization.

\subsection{Routing configurations}
\label{app:routing-configs}

We report the routing-layer configurations used in our experiments. LLaVA-1.5 contains 32 decoder layers and uses 576 visual tokens under the standard evaluation setting. Table~\ref{tab:llava-routing-config} summarizes the LLaVA-1.5 configurations used in the main \mbox{Avg. = 192} comparison.

\begin{table}[t]
\centering
\footnotesize
\setlength{\tabcolsep}{4.0pt}
\renewcommand{\arraystretch}{0.95}
\caption{LLaVA-1.5 routing configurations at \mbox{Avg. = 192}. All schedules use keep-index positional handling.}
\label{tab:llava-routing-config}
\begin{tabular}{@{}llcl@{}}
\toprule
\textbf{Backbone} & \textbf{Setting} & \textbf{Layers} & \textbf{Routing layers} \\
\midrule
LLaVA-1.5-7B & PDrop (L8)           & 32 & $\{8,16,24\}$ \\
LLaVA-1.5-7B & PDrop (L2)           & 32 & $\{2,7,15,23\}$ \\
LLaVA-1.5-7B & PDrop (L2) + Reroute & 32 & $\{2,7,15,23\}$ \\
LLaVA-1.5-7B & FastV (L3)           & 32 & $\{3\}$ \\
LLaVA-1.5-7B & FastV (L3) + Reroute & 32 & $\{3,7,15,23\}$ \\
\bottomrule
\end{tabular}
\end{table}

PDrop introduces a configuration ambiguity because different routing-layer schedules can target the same average-token budget. We consider two conventions: a standard PDrop-style schedule that distributes routing decisions approximately evenly across decoder depth, e.g., $\{8,16,24\}$ for LLaVA-1.5, and an early-routing convention that starts near layer 2 and continues at later stages, e.g., $\{2,7,15,23\}$. We denote these configurations as PDrop (L8) and PDrop (L2), respectively. Appendix~\ref{app:pdrop-schedule-ablation} shows that PDrop (L2) gives stronger grounding results, so we use the early-routing convention as the main PDrop-aligned schedule for Reroute.

For cross-model experiments, we preserve the same scheduling principle while normalizing routing layers to model depth and respecting architecture-specific constraints. Table~\ref{tab:cross-model-routing} summarizes the cross-model routing-layer conventions used for Reroute.

\begin{table*}[t]
\centering
\small
\setlength{\tabcolsep}{4.5pt}
\renewcommand{\arraystretch}{1.0}
\caption{Cross-model routing-layer conventions used for Reroute.}
\label{tab:cross-model-routing}
\resizebox{0.98\textwidth}{!}{%
\begin{tabular}{@{}llcll@{}}
\toprule
\textbf{Backbone} & \textbf{Setting} & \textbf{Layers} & \textbf{Routing layers} & \textbf{Rationale} \\
\midrule
LLaVA-1.5-7B  & PDrop + Reroute & 32 & $\{2,7,15,23\}$ & Early PDrop-aligned schedule from the LLaVA ablation \\
LLaVA-1.5-7B  & FastV + Reroute & 32 & $\{3,7,15,23\}$ & Main-table FastV setting uses $K=3$ \\
Qwen2.5-VL-7B & PDrop + Reroute & 28 & $\{2,6,13,20\}$ & Layer-normalized early PDrop-aligned schedule \\
Qwen2.5-VL-7B & FastV + Reroute & 28 & $\{3,6,13,20\}$ & Layer-normalized FastV-aligned schedule \\
Qwen3-VL-8B   & Both             & 36 & $\{3,8,17,26\}$ & Starts after DeepStack injection layers \\
Qwen3.5-9B    & Both             & 32 & $\{3,7,15,23\}$ & Snapped to compatible full-attention layers in the hybrid-attention backbone \\
\bottomrule
\end{tabular}%
}
\end{table*}

\subsection{Position-index handling for physical pruning}
\label{app:position-index-handling}

When visual tokens are physically pruned, the remaining tokens can either be re-indexed into a contiguous position sequence or keep their original visual position indices~\cite{chien2025groundingawaretokenpruningrecovering,huang2026n,bagrov2025efficientvideosamplingpruning}. Table~\ref{tab:reindex-vs-gap} compares these two choices under FastV on LLaVA-1.5-7B. Re-indexing substantially degrades both general VQA and RefCOCO grounding, indicating that spatial position preservation is critical for localization after token removal.

Therefore, for all physical-pruning baselines, we use the keep-index convention: kept visual tokens preserve their original position indices after pruning. This makes the pruning baselines stronger and avoids conflating pruning failures with position-indexing artifacts. Reroute does not physically remove deferred tokens, so it preserves the original sequence layout by construction.

\begin{table}[t]
\centering
\small
\setlength{\tabcolsep}{4.5pt}
\renewcommand{\arraystretch}{0.95}
\caption{Re-indexing vs. preserved indexing under FastV on LLaVA-1.5-7B. Grounding results are reported using ACC@0.5.}
\label{tab:reindex-vs-gap}
\begin{tabular}{@{}lclcccc@{}}
\toprule
\textbf{Schedule} & \textbf{Avg.} & \textbf{PE handling} & \textbf{GQA} & \textbf{MMBench} & \textbf{RefCOCO A} & \textbf{RefCOCO B} \\
\midrule
FastV (K=2) & 306 & Re-index   & 5.3  & 16.6 & 3.9  & 3.3  \\
FastV (K=2) & 306 & Keep-index & 58.9 & 61.9 & 45.0 & 37.0 \\
FastV (K=3) & 192 & Re-index   & 40.8 & 34.8 & 1.9  & 1.1  \\
FastV (K=3) & 192 & Keep-index & 55.0 & 58.6 & 27.8 & 22.3 \\
\bottomrule
\end{tabular}
\end{table}

\section{Routing Schedule Analysis}
\label{app:schedule-analysis}

This section analyzes the schedule-design choices behind the main Reroute configurations. Since Reroute is training-free and reuses the same attention-score ranking interface as pruning baselines, its main design degrees of freedom are the routing layers and per-stage retention ratios. We therefore study three related questions: whether PDrop should start early or follow its standard depth-spaced schedule, how many reroute stages are useful, and where these stages should be placed across decoder depth.

\subsection{Early-stage routing for PDrop}
\label{app:pdrop-schedule-ablation}

For the full RefCOCO-family evaluation, we report ACC@0.5 as the primary grounding metric in Table~\ref{tab:pdrop-schedule-refcoco-acc05}. Across both evaluated backbones, PDrop (L2) consistently improves over PDrop (L8) at the same \mbox{Avg. = 192} budget. This result supports using early-stage routing as the main PDrop-aligned schedule when comparing pruning against Reroute.

\begin{table*}[t]
\centering
\small
\setlength{\tabcolsep}{4.0pt}
\renewcommand{\arraystretch}{0.95}
\setlength{\aboverulesep}{0.20ex}
\setlength{\belowrulesep}{0.20ex}
\setlength{\cmidrulesep}{0.12ex}
\caption{Schedule ablation on RefCOCO-family visual grounding at \mbox{Avg. = 192}. All values report ACC@0.5. We compare the standard PDrop schedule, denoted PDrop (L8), with the early-routing schedule, denoted PDrop (L2), on the two backbones used for the main grounding analysis.}
\label{tab:pdrop-schedule-refcoco-acc05}
\resizebox{0.96\textwidth}{!}{%
\begin{tabular}{@{}llc*{8}{c}@{}}
\toprule
\textbf{Model} & \textbf{Schedule} & \textbf{Avg.}
& \multicolumn{3}{c}{\textbf{RefCOCO}}
& \multicolumn{3}{c}{\textbf{RefCOCO+}}
& \multicolumn{2}{c}{\textbf{RefCOCOg}} \\
\cmidrule(lr){4-6}
\cmidrule(lr){7-9}
\cmidrule(lr){10-11}
& & & \textbf{val} & \textbf{A} & \textbf{B}
& \textbf{val} & \textbf{A} & \textbf{B}
& \textbf{val} & \textbf{test} \\
\midrule
\rowcolor{baserow}
LLaVA-1.5-7B & PDrop (L8) & 192
& 26.2 & 31.9 & 19.9
& 22.4 & 27.1 & 16.5
& 21.8 & 20.8 \\
\rowcolor{baserow}
LLaVA-1.5-7B & PDrop (L2) & 192
& 39.8 & 45.1 & 33.4
& 33.5 & 39.5 & 25.9
& 33.2 & 32.8 \\
\midrule
\rowcolor{baserow}
Qwen2.5-VL-7B & PDrop (L8) & 192
& 33.4 & 39.3 & 28.5
& 28.9 & 34.6 & 23.3
& 32.7 & 33.1 \\
\rowcolor{baserow}
Qwen2.5-VL-7B & PDrop (L2) & 192
& 55.9 & 59.2 & 52.0
& 48.4 & 53.0 & 43.6
& 53.6 & 52.9 \\
\bottomrule
\end{tabular}%
}
\end{table*}

\subsection{Frequency and location sweeps for Reroute}
\label{app:reroute-schedule-ablations}

We provide additional details for the routing-schedule ablations used in the main paper. Table~\ref{tab:reroute-frequency-config} lists the exact schedules used for the frequency sweep in Figure~\ref{fig:ablation-freq}, where the first routing layer is fixed and the number of reroute stages is varied. Table~\ref{tab:reroute-location-config} reports a complementary location sweep, where the number of reroute stages is fixed but their decoder-layer positions are changed. In both sweeps, each routing stage uses the same per-stage retention ratio, $r_i=0.5$. The realized average token ratio is reported explicitly because it can vary with the placement of routing layers.

\begin{table*}[t]
\centering
\small
\caption{\textbf{Frequency-sweep configurations for reroute scheduling on LLaVA-1.5-7B.} All settings fix the first routing layer at layer 2 and use per-stage retention ratio $r_i=0.5$. The sweep varies only the number of reroute stages. RefCOCO A/B denote testA/testB Acc@0.5.}
\label{tab:reroute-frequency-config}
\setlength{\tabcolsep}{4.5pt}
\renewcommand{\arraystretch}{1.05}
\begin{tabularx}{\textwidth}{@{}l c X c c c c@{}}
\toprule
\textbf{Setting}
& \textbf{\#Stages}
& \textbf{Routing layers}
& \textbf{Avg. ratio}
& \textbf{GQA}
& \textbf{RefCOCO A}
& \textbf{RefCOCO B} \\
\midrule
N1 & 1 & $\{2\}$ & 0.5312 & 58.94 & 44.95 & 36.90 \\
N2 & 2 & $\{2,16\}$ & 0.5312 & 59.03 & 44.56 & 36.31 \\
N3 & 3 & $\{2,11,20\}$ & 0.5312 & 59.24 & 45.48 & 37.49 \\
N4 & 4 & $\{2,9,16,23\}$ & 0.5312 & 59.56 & 48.49 & 37.45 \\
N6 & 6 & $\{2,7,12,17,22,27\}$ & 0.5312 & 59.72 & 50.93 & 39.33 \\
N8 & 8 & $\{2,6,10,14,18,22,26,30\}$ & 0.5312 & 59.79 & 52.22 & 41.39 \\
Dense & 30 & $\{2,3,\ldots,31\}$ & 0.5312 & 59.88 & 53.51 & 41.18 \\
\bottomrule
\end{tabularx}
\end{table*}

\begin{table*}[t]
\centering
\small
\caption{\textbf{Location-sweep configurations for reroute scheduling on LLaVA-1.5-7B.} All settings use four reroute stages with per-stage retention ratio $r_i=0.5$, and vary where these stages are placed across decoder depth. RefCOCO A/B denote testA/testB Acc@0.5.}
\label{tab:reroute-location-config}
\setlength{\tabcolsep}{5.0pt}
\renewcommand{\arraystretch}{1.05}
\begin{tabularx}{\textwidth}{@{}l X c c c c@{}}
\toprule
\textbf{Placement}
& \textbf{Routing layers}
& \textbf{Avg. ratio}
& \textbf{GQA}
& \textbf{RefCOCO A}
& \textbf{RefCOCO B} \\
\midrule
Early & $\{2,5,9,13\}$ & 0.531 & 60.05 & 53.07 & 41.28 \\
Uniform & $\{4,12,20,28\}$ & 0.563 & 59.31 & 46.88 & 37.74 \\
Late & $\{9,15,21,27\}$ & 0.641 & 60.61 & 53.81 & 40.29 \\
Very late & $\{15,19,23,27\}$ & 0.734 & 60.53 & 59.41 & 47.93 \\
Clustered early & $\{2,4,7,11\}$ & 0.531 & 59.89 & 53.08 & 41.06 \\
Clustered late & $\{20,23,26,29\}$ & 0.813 & 60.52 & 59.02 & 47.91 \\
\bottomrule
\end{tabularx}
\end{table*}

The frequency sweep shows that increasing the number of reroute decisions generally improves both GQA and RefCOCO grounding while keeping the average token ratio fixed. The location sweep further shows that the placement of reroute stages matters: later routing stages retain a larger realized average token ratio and perform strongly on RefCOCO, while early and clustered-early schedules offer a stricter-budget comparison. Together, these ablations indicate that Reroute benefits from repeated re-evaluation, but its schedule should be chosen with both routing frequency and decoder-depth placement in mind.

\section{Supplementary Results and Visualizations}
\label{app:supp-results}

This section provides additional evidence beyond the main grounding tables. We first report general VQA results to verify that Reroute does not sacrifice broad multimodal capability. We then discuss Qwen3-VL as an architecture-sensitive case study, followed by qualitative grounding and token-selection visualizations.

\subsection{General VQA preservation}
\label{app:general-vqa}

We report general VQA benchmark results in Tables~\ref{tab:llava15_vqa_original_format}, \ref{tab:llava15_vqa_hf_format}, \ref{tab:qwen25vl_vqa_main}, \ref{tab:qwen3vl_vqa_main}, and \ref{tab:qwen35_vqa_main}. These results serve as a preservation test rather than the primary source of improvement. Since many VQA benchmarks can be answered from coarse or globally salient visual evidence, the key question is whether Reroute maintains broad multimodal capability under the same average-token budget.

Across LLaVA-1.5-7B and Qwen-family backbones, Reroute is generally comparable to, and often improves over, its corresponding pruning baseline. The trend is clearest for PDrop + Reroute, where multi-stage routing gives deferred tokens later opportunities to re-enter computation; for example, on Qwen2.5-VL-7B, the average-ratio score improves over PDrop at all three reduction levels. FastV + Reroute is less stable under extreme reduction because FastV depends on a single early scoring layer, but remains broadly comparable under moderate budgets. Overall, these results indicate that Reroute's grounding gains do not come from sacrificing general VQA performance. Instead, recoverable routing provides a robust alternative to irreversible pruning, especially when paired with multi-stage schedules.


\begin{table*}[t]
\centering
\caption{
VQA performance comparison on LLaVA-1.5-7B under the original LLaVA / N\"uwa
official-codebase format. Prior-method rows are paper-reported results, while
rows marked with \NuwaOfficial{} are reproduced in the official N\"uwa codebase.
Only tasks available in the updated result sheet are reported.
Best and second-best results within each average-token block are shown in
\best{bold} and \second{underlined}, respectively.
}
\label{tab:llava15_vqa_original_format}
\scriptsize
\setlength{\tabcolsep}{3.0pt}
\renewcommand{\arraystretch}{1.08}
\resizebox{\linewidth}{!}{%
\begin{tabular}{@{}llccccccccc@{}}
\toprule
\textbf{Method} & \textbf{Source}
& \textbf{GQA} & \textbf{MMB} & \textbf{MMMU} & \textbf{MME}
& \textbf{TextVQA} & \textbf{POPE} & \textbf{SQA}
& \textbf{SEED} & \textbf{Avg. Ratio} \\
\midrule

\multicolumn{11}{c}{\textit{Average Token 576}} \\
\midrule

\rowcolor{vanillarow}
Vanilla
& CVPR'24
& \best{61.9} & \best{64.7} & \best{36.3} & \best{1862}
& \best{58.2} & \best{85.9} & \best{69.5}
& \best{58.6} & \best{100.0\%} \\

\midrule
\multicolumn{11}{c}{\textit{Average Token 192} $\downarrow 66.7\%$} \\
\midrule

\rowcolor{otherrow}
FastV
& ECCV'24
& 52.7 & 61.2 & 34.3 & 1612
& 52.5 & 64.8 & 67.3
& 57.1 & 90.1\% \\

\rowcolor{otherrow}
PDrop
& CVPR'25
& 57.1 & 63.2 & 34.1 & 1766
& 56.1 & 82.3 & \best{70.2}
& 54.7 & 95.7\% \\

\rowcolor{otherrow}
SparseVLM
& ICML'25
& 57.6 & 62.5 & 33.8 & 1721
& 56.1 & 83.6 & \second{69.1}
& 55.8 & 95.4\% \\

\rowcolor{otherrow}
VisionZip
& CVPR'25
& 59.3 & 63.0 & \best{36.6} & 1782
& 57.3 & 85.3 & 68.9
& 56.4 & 97.9\% \\

\midrule

\rowcolor{baserow}
N\"uwa
& ICLR'26
& \second{60.9} & \best{64.3} & 35.5 & \second{1834}
& \second{57.4} & \best{86.4} & 68.2
& \second{59.7} & \second{99.2\%} \\

\rowcolor{routerow}
N\"uwa+Reroute\NuwaOfficial
& Ours
& \best{61.1} & \second{63.6} & \second{35.7} & \best{1848}
& \best{57.8} & \second{86.0} & 68.6
& \best{66.1} & \best{100.6\%} \\

\midrule
\multicolumn{11}{c}{\textit{Average Token 128} $\downarrow 77.8\%$} \\
\midrule

\rowcolor{otherrow}
FastV
& ECCV'24
& 49.6 & 56.1 & 34.9 & 1490
& 50.6 & 59.6 & 60.2
& 55.9 & 85.2\% \\

\rowcolor{otherrow}
PDrop
& CVPR'25
& 56.0 & 61.1 & 34.2 & 1664
& 55.1 & 82.3 & \best{69.9}
& 53.3 & 93.8\% \\

\rowcolor{otherrow}
SparseVLM
& ICML'25
& 56.0 & 60.0 & 33.8 & 1696
& 54.9 & 80.5 & 67.1
& 53.4 & 92.9\% \\

\rowcolor{otherrow}
VisionZip
& CVPR'25
& 57.6 & 62.0 & \best{37.9} & 1761
& 56.8 & 83.2 & \second{68.9}
& 54.9 & 96.9\% \\

\rowcolor{otherrow}
PruMerge
& ICCV'25
& 57.8 & 59.6 & \second{36.2} & 1712
& 54.3 & 81.5 & 67.6
& \na & 94.7\% \\

\midrule

\rowcolor{baserow}
N\"uwa
& ICLR'26
& \second{60.2} & \best{63.4} & 35.8 & \best{1828}
& \second{57.0} & \best{85.5} & 67.8
& \second{58.7} & \second{98.4\%} \\

\rowcolor{routerow}
N\"uwa+Reroute\NuwaOfficial
& Ours
& \best{60.4} & \second{63.2} & 36.1 & \second{1827}
& \best{57.3} & \second{85.3} & 68.2
& \best{64.4} & \best{99.8\%} \\

\midrule
\multicolumn{11}{c}{\textit{Average Token 64} $\downarrow 88.9\%$} \\
\midrule

\rowcolor{otherrow}
FastV
& ECCV'24
& 46.1 & 48.0 & 34.0 & 1256
& 47.8 & 59.6 & 51.1
& 51.9 & 77.9\% \\

\rowcolor{otherrow}
PDrop
& CVPR'25
& 41.9 & 33.3 & 26.5 & 1092
& 45.9 & 55.9 & 69.2
& 40.0 & 70.3\% \\

\rowcolor{otherrow}
SparseVLM
& ICML'25
& 53.8 & 60.1 & 35.4 & 1589
& 53.4 & 77.5 & \best{69.8}
& 51.1 & 91.5\% \\

\rowcolor{otherrow}
VisionZip
& CVPR'25
& 55.1 & 60.1 & \second{36.2} & \second{1690}
& \best{55.5} & 77.0 & 69.0
& 52.2 & 93.2\% \\

\rowcolor{otherrow}
PruMerge
& ICCV'25
& 55.4 & 59.6 & 35.8 & 1616
& 52.0 & 75.7 & \second{69.5}
& \na & 92.1\% \\

\midrule

\rowcolor{baserow}
N\"uwa
& ICLR'26
& \second{58.3} & \best{62.0} & \best{36.4} & \best{1706}
& 54.9 & \best{83.0} & 67.5
& \second{56.4} & \second{95.8\%} \\

\rowcolor{routerow}
N\"uwa+Reroute\NuwaOfficial
& Ours
& \best{58.5} & \second{61.9} & 35.9 & 1688
& \second{55.1} & \second{82.6} & 67.7
& \best{61.6} & \best{96.7\%} \\

\bottomrule
\end{tabular}%
}

\vspace{0.35em}
\footnotesize{
\colorbox{vanillarow}{Vanilla baseline};
\colorbox{otherrow}{Other methods};
\colorbox{baserow}{Base methods};
\colorbox{routerow}{Reroute variants}. \\
\NuwaOfficial{}: reproduced in the official N\"uwa codebase with Reroute applied.
Unavailable results are denoted by ``--''.
MMB denotes MMBench; SQA denotes ScienceQA; SEED denotes SeedBench.
Avg. Ratio is reported relative to the corresponding full-token baseline.
}
\end{table*}


\begin{table*}[t]
\centering
\caption{
VQA performance comparison on Qwen2.5-VL-7B. Best and second-best results
within each average-token block are shown in \best{bold} and
\second{underlined}, respectively.
}
\label{tab:qwen25vl_vqa_main}
\scriptsize
\setlength{\tabcolsep}{2.7pt}
\renewcommand{\arraystretch}{1.08}
\resizebox{\linewidth}{!}{%
\begin{tabular}{llcccccccccccc}
\toprule
\textbf{Method} & \textbf{Source}
& \textbf{GQA} & \textbf{POPE} & \textbf{MMB} & \textbf{SQA}
& \textbf{TextVQA} & \textbf{AI2D} & \textbf{MMMU}
& \textbf{MMStar} & \textbf{SEED} & \textbf{VQAv2-lite}
& \textbf{MME} & \textbf{Avg. Ratio} \\
\midrule

\multicolumn{14}{c}{\textit{Average Token (100\%)}} \\
\midrule
\rowcolor{vanillarow}
Vanilla
& Qwen'25
& \best{60.8} & \best{87.8} & \best{83.7} & \best{87.8}
& \best{81.0} & \best{82.6} & \best{48.7}
& \best{62.3} & \best{79.8} & \best{78.6}
& \best{2318} & \best{100.0\%} \\

\midrule
\multicolumn{14}{c}{\textit{Average Token Reduction} $\downarrow 66.7\%$} \\
\midrule
\rowcolor{baserow}
FastV
& ECCV'24
& 54.0 & 84.6 & 77.8 & 82.8
& 80.1 & 72.5 & \best{49.6}
& 50.3 & 66.9 & 72.9
& 2178 & 92.0\% \\

\rowcolor{baserow}
PDrop
& CVPR'25
& 55.7 & 84.7 & \second{78.6} & \best{85.0}
& \second{81.3} & \best{75.0} & \second{48.4}
& \second{52.6} & \second{68.7} & 75.8
& \second{2213} & \second{93.8\%} \\

\midrule
\rowcolor{routerow}
FastV+Reroute
& Ours
& \second{55.9} & \second{85.2} & 77.3 & 82.5
& 77.9 & 73.2 & 47.3
& 50.6 & 68.1 & \second{76.3}
& 2184 & 92.3\% \\

\rowcolor{routerow}
PDrop+Reroute
& Ours
& \best{57.0} & \best{85.8} & \best{79.0} & \second{84.8}
& \best{82.2} & \second{74.8} & 46.7
& \best{53.5} & \best{69.5} & \best{77.6}
& \best{2243} & \best{94.4\%} \\

\midrule
\multicolumn{14}{c}{\textit{Average Token Reduction} $\downarrow 77.8\%$} \\
\midrule
\rowcolor{baserow}
FastV
& ECCV'24
& 48.2 & 79.5 & 66.5 & 77.1
& 74.2 & 67.5 & 44.9
& 41.0 & 61.0 & 64.2
& 1994 & 83.0\% \\

\rowcolor{baserow}
PDrop
& CVPR'25
& 50.7 & 80.5 & \second{72.8} & \best{82.0}
& \second{77.5} & \best{71.5} & \best{47.0}
& \second{46.8} & \second{62.0} & 67.6
& \second{2056} & \second{87.4\%} \\

\midrule
\rowcolor{routerow}
FastV+Reroute
& Ours
& \second{52.1} & \second{80.8} & 68.1 & 78.8
& 72.1 & 68.1 & 45.0
& 42.4 & 59.4 & \second{72.4}
& 2002 & 84.9\% \\

\rowcolor{routerow}
PDrop+Reroute
& Ours
& \best{53.8} & \best{83.1} & \best{75.4} & \second{81.2}
& \best{80.3} & \second{71.1} & \second{45.8}
& \best{47.5} & \best{62.6} & \best{74.1}
& \best{2078} & \best{89.4\%} \\

\midrule
\multicolumn{14}{c}{\textit{Average Token Reduction} $\downarrow 88.9\%$} \\
\midrule
\rowcolor{baserow}
FastV
& ECCV'24
& 36.3 & 49.3 & 32.0 & 73.2
& 13.6 & 64.7 & 41.2
& 30.6 & 43.0 & 37.7
& 1127 & 56.1\% \\

\rowcolor{baserow}
PDrop
& CVPR'25
& \second{44.9} & \second{69.3} & \best{67.4} & \best{79.4}
& \second{53.5} & \best{68.0} & \best{45.4}
& \second{39.7} & \second{53.7} & \second{58.2}
& \second{1755} & \second{76.9\%} \\

\midrule
\rowcolor{routerow}
FastV+Reroute
& Ours
& 42.2 & 58.3 & 36.3 & 74.1
& 36.7 & 64.3 & 41.7
& 30.8 & 45.3 & 52.3
& 1247 & 63.5\% \\

\rowcolor{routerow}
PDrop+Reroute
& Ours
& \best{49.3} & \best{77.0} & \second{66.0} & \second{77.2}
& \best{65.0} & \second{67.8} & \second{44.4}
& \best{41.0} & \best{57.2} & \best{68.6}
& \best{1874} & \best{81.3\%} \\

\bottomrule
\end{tabular}
}

\vspace{0.3em}
\footnotesize{
\colorbox{vanillarow}{Vanilla baseline};
\colorbox{baserow}{Base methods};
\colorbox{routerow}{Reroute variants}.\\
Avg. Ratio is the mean score ratio against the vanilla baseline,
computed over subsets where both scores are available.
Schedule details are reported in the appendix.
}
\end{table*}



\begin{table*}[t]
\centering
\caption{
VQA performance comparison on Qwen3-VL-8B.
Best and second-best results within each average-token block are shown in
\best{bold} and \second{underlined}, respectively.
}
\label{tab:qwen3vl_vqa_main}
\scriptsize
\setlength{\tabcolsep}{2.7pt}
\renewcommand{\arraystretch}{1.08}
\resizebox{\linewidth}{!}{%
\begin{tabular}{llcccccccccccc}
\toprule
\textbf{Method} & \textbf{Source}
& \textbf{GQA} & \textbf{POPE} & \textbf{MMB} & \textbf{SQA}
& \textbf{TextVQA} & \textbf{AI2D} & \textbf{MMMU}
& \textbf{MMStar} & \textbf{SEED} & \textbf{VQAv2-lite}
& \textbf{MME} & \textbf{Avg. Ratio} \\
\midrule

\multicolumn{14}{c}{\textit{Average Token (100\%)}} \\
\midrule
\rowcolor{vanillarow}
Vanilla
& Qwen'25
& \best{61.7} & \best{88.9} & \best{84.4} & \best{94.5}
& \best{81.6} & \best{83.8} & \best{52.2}
& \best{61.8} & \best{80.4} & \best{80.0}
& \best{2374} & \best{100.0\%} \\

\midrule
\multicolumn{14}{c}{\textit{Average Token Reduction} $\downarrow 66.7\%$} \\
\midrule
\rowcolor{baserow}
FastV
& ECCV'24
& 54.1 & 85.4 & \best{76.6} & \best{87.3}
& \second{76.7} & \best{74.8} & 49.2
& \second{49.1} & 72.7 & 70.6
& \second{2012} & 89.8\% \\

\rowcolor{baserow}
PDrop 
& CVPR'25
& \second{56.9} & \best{87.6} & 75.4 & \second{87.0}
& 71.0 & 74.4 & \best{50.3}
& \best{50.1} & \best{77.6} & \second{74.2}
& 2008 & \second{90.8\%} \\

\midrule
\rowcolor{routerow}
FastV+Reroute
& Ours
& 54.7 & 85.4 & 71.3 & 81.2
& 73.0 & 71.0 & 48.3
& 43.7 & 71.7 & 71.8
& 1902 & 86.5\% \\

\rowcolor{routerow}
PDrop+Reroute
& Ours
& \best{57.5} & \second{87.1} & \second{75.8} & 86.9
& \best{77.8} & \second{74.6} & \second{49.6}
& 48.4 & \second{76.8} & \best{75.1}
& \best{2039} & \best{91.4\%} \\

\midrule
\multicolumn{14}{c}{\textit{Average Token Reduction} $\downarrow 77.8\%$} \\
\midrule
\rowcolor{baserow}
FastV
& ECCV'24
& 49.4 & 80.7 & \second{68.0} & \best{83.7}
& \second{68.5} & \second{69.5} & \second{48.8}
& \second{43.0} & 69.3 & \second{63.4}
& 1756 & \second{82.7\%} \\

\rowcolor{baserow}
PDrop
& CVPR'25
& \second{51.7} & \second{83.5} & 67.9 & 80.8
& 57.6 & 68.9 & 47.9
& \best{43.6} & \second{70.7} & 62.5
& \second{1812} & 81.9\% \\

\midrule
\rowcolor{routerow}
FastV+Reroute
& Ours
& 50.4 & 79.1 & 62.7 & 77.3
& 64.2 & 67.2 & 48.0
& 40.0 & 64.0 & 62.1
& 1659 & 79.0\% \\

\rowcolor{routerow}
PDrop+Reroute
& Ours
& \best{53.6} & \best{84.2} & \best{70.3} & \second{81.6}
& \best{71.7} & \best{70.2} & \best{49.1}
& 42.5 & \best{71.9} & \best{69.3}
& \best{1839} & \best{85.4\%} \\

\midrule
\multicolumn{14}{c}{\textit{Average Token Reduction} $\downarrow 88.9\%$} \\
\midrule
\rowcolor{baserow}
FastV
& ECCV'24
& 37.3 & 56.9 & 29.6 & 73.8
& 16.0 & 64.9 & 44.9
& 30.7 & 48.9 & 40.8
& 1195 & 57.5\% \\

\rowcolor{baserow}
PDrop
& CVPR'25
& \second{45.3} & \second{73.8} & \second{52.6} & \second{75.5}
& \second{33.5} & \second{65.3} & \second{46.8}
& \second{35.4} & \second{58.4} & \second{51.6}
& \second{1525} & \second{69.6\%} \\

\midrule
\rowcolor{routerow}
FastV+Reroute
& Ours
& 39.8 & 57.9 & 29.0 & 73.3
& 20.0 & 64.5 & 44.1
& 30.1 & 47.9 & 42.7
& 1177 & 58.1\% \\

\rowcolor{routerow}
PDrop+Reroute
& Ours
& \best{49.3} & \best{77.5} & \best{60.0} & \best{77.1}
& \best{60.6} & \best{66.8} & \best{46.9}
& \best{38.6} & \best{63.8} & \best{60.6}
& \best{1661} & \best{77.4\%} \\

\bottomrule
\end{tabular}%
}
\vspace{0.3em}
\footnotesize{
\colorbox{vanillarow}{Vanilla baseline};
\colorbox{baserow}{Base methods};
\colorbox{routerow}{Reroute variants}.\\
Avg. Ratio is the mean score ratio against the vanilla baseline,
computed over subsets where both scores are available.
Schedule details are reported in the appendix.
}
\end{table*}


\begin{table*}[t]
\centering
\caption{
VQA performance comparison on Qwen3.5-9B-Hybrid.
Best and second-best results within each average-token block are shown in
\best{bold} and \second{underlined}, respectively.
}
\label{tab:qwen35_vqa_main}
\scriptsize
\setlength{\tabcolsep}{2.7pt}
\renewcommand{\arraystretch}{1.08}
\resizebox{\linewidth}{!}{%
\begin{tabular}{llcccccccccccc}
\toprule
\textbf{Method} & \textbf{Source}
& \textbf{GQA} & \textbf{POPE} & \textbf{MMB} & \textbf{SQA}
& \textbf{TextVQA} & \textbf{AI2D} & \textbf{MMMU}
& \textbf{MMStar} & \textbf{SEED} & \textbf{VQAv2-lite}
& \textbf{MME} & \textbf{Avg. Ratio} \\
\midrule

\multicolumn{14}{c}{\textit{Average Gated Attention Token (100\%)}} \\
\midrule
\rowcolor{vanillarow}
Vanilla
& Qwen'25
& \best{61.7} & \best{89.6} & \best{84.0} & \best{93.5}
& \best{82.8} & \best{85.2} & \best{48.1}
& \best{46.8} & \best{77.0} & \best{79.6}
& \best{2378} & \best{100.0\%} \\

\midrule
\multicolumn{14}{c}{\textit{Average Gated Attention Token Reduction} $\downarrow 66.7\%$} \\
\midrule
\rowcolor{baserow}
FastV
& ECCV'24
& 52.8 & 81.4 & 69.9 & 82.8
& 48.0 & 72.0 & \second{45.3}
& 35.4 & 70.1 & 62.8
& 1823 & 82.5\% \\

\rowcolor{baserow}
PDrop
& CVPR'25
& 56.6 & \second{85.5} & 74.2 & 86.7
& 65.0 & 74.6 & 44.3
& 37.4 & \second{72.1} & 68.6
& 1924 & 87.9\% \\

\midrule
\rowcolor{routerow}
FastV+Reroute
& Ours
& \second{58.4} & 84.7 & \second{76.9} & \second{87.2}
& \second{77.4} & \second{75.3} & 44.3
& \second{38.9} & 71.1 & \best{77.2}
& \second{1929} & \second{91.1\%} \\

\rowcolor{routerow}
PDrop+Reroute
& Ours
& \best{59.6} & \best{87.2} & \best{80.0} & \best{89.9}
& \best{79.6} & \best{77.3} & \best{48.1}
& \best{40.0} & \best{72.7} & \second{75.4}
& \best{2042} & \best{93.9\%} \\

\midrule
\multicolumn{14}{c}{\textit{Average Gated Attention Token Reduction} $\downarrow 77.8\%$} \\
\midrule
\rowcolor{baserow}
FastV
& ECCV'24
& 47.5 & 74.7 & 58.9 & 78.3
& 28.0 & 69.2 & 43.2
& 31.2 & 61.8 & 53.1
& 1547 & 72.5\% \\

\rowcolor{baserow}
PDrop
& CVPR'25
& 51.7 & 80.1 & 65.7 & 79.3
& 46.2 & 71.4 & 44.3
& 33.2 & 64.9 & 58.6
& 1688 & 78.9\% \\

\midrule
\rowcolor{routerow}
FastV+Reroute
& Ours
& \second{56.4} & \second{82.5} & \second{72.6} & \second{82.0}
& \second{71.7} & \second{72.2} & \best{45.0}
& \second{35.1} & \second{65.1} & \second{72.3}
& \second{1798} & \second{86.2\%} \\

\rowcolor{routerow}
PDrop+Reroute
& Ours
& \best{57.3} & \best{83.9} & \best{75.6} & \best{84.9}
& \best{74.6} & \best{74.3} & \second{44.6}
& \best{37.7} & \best{71.7} & \best{74.4}
& \best{1932} & \best{89.6\%} \\

\midrule
\multicolumn{14}{c}{\textit{Average Gated Attention Token Reduction} $\downarrow 88.9\%$} \\
\midrule
\rowcolor{baserow}
FastV
& ECCV'24
& 38.4 & 55.7 & 33.2 & 73.1
& 9.8 & 66.8 & \second{43.3}
& 20.1 & 46.9 & 38.9
& 1263 & 57.1\% \\

\rowcolor{baserow}
PDrop
& CVPR'25
& 44.8 & \second{69.0} & 51.7 & 75.3
& 24.3 & \second{68.9} & 42.3
& 27.7 & 56.6 & 47.3
& 1431 & 67.5\% \\

\midrule
\rowcolor{routerow}
FastV+Reroute
& Ours
& \second{51.0} & 68.7 & \second{64.0} & \second{78.5}
& \second{56.0} & 67.3 & 42.8
& \second{29.5} & \second{56.8} & \second{64.5}
& \second{1539} & \second{76.1\%} \\

\rowcolor{routerow}
PDrop+Reroute
& Ours
& \best{54.9} & \best{79.7} & \best{70.8} & \best{80.8}
& \best{68.3} & \best{71.9} & \best{45.1}
& \best{34.7} & \best{64.9} & \best{70.8}
& \best{1711} & \best{84.4\%} \\

\bottomrule
\end{tabular}%
}
\vspace{0.3em}
\footnotesize{
\colorbox{vanillarow}{Vanilla baseline};
\colorbox{baserow}{Base methods};
\colorbox{routerow}{Reroute variants}.\\
Avg. Ratio is the mean score ratio against the vanilla baseline,
computed over subsets where both scores are available.
Schedule details are reported in the appendix.
}
\end{table*}

\subsection{Qwen3-VL grounding case study}
\label{app:qwen3vl-case-study}

Table~\ref{tab:qwen3vl_refcoco_main} reports the Qwen3-VL RefCOCO results. We interpret this experiment as an architecture-sensitive diagnostic rather than as a primary success case. Reroute is training-free and does not learn a router; it reuses the same attention-score ranking signal as the corresponding pruning baseline. Therefore, the relevant design variable is the hand-specified routing schedule, including routed layers and keep-ratio budgets, rather than a trained routing policy.



\begin{table}[t]
\centering
\caption{
Performance comparison on RefCOCO-series visual grounding benchmarks using
Qwen3-VL-8B. Best and second-best results within each average-token block are
shown in \best{bold} and \second{underlined}, respectively.
}
\label{tab:qwen3vl_refcoco_main}
\setlength{\tabcolsep}{3.0pt}
\renewcommand{\arraystretch}{1.08}
\resizebox{\linewidth}{!}{%
\begin{tabular}{@{}llccccccccc@{}}
\toprule
\textbf{Method} & \textbf{Source}
& \multicolumn{3}{c}{\textbf{RefCOCO}}
& \multicolumn{3}{c}{\textbf{RefCOCO+}}
& \multicolumn{2}{c}{\textbf{RefCOCOg}}
& \shortstack[c]{\textbf{Avg.}} \\
\cmidrule(lr){3-5}
\cmidrule(lr){6-8}
\cmidrule(lr){9-10}
& & \textbf{val} & \textbf{A} & \textbf{B}
& \textbf{val} & \textbf{A} & \textbf{B}
& \textbf{val} & \textbf{test}
& \textbf{Ratio}\\
\midrule

\multicolumn{11}{c}{\textit{Average Token (100\%)}} \\
\midrule
\rowcolor{vanillarow}
Vanilla
& Qwen'25
& \best{89.3} & \best{90.4} & \best{84.7}
& \best{83.3} & \best{85.0} & \best{76.3}
& \best{87.2} & \best{87.2}
& \best{100.0\%} \\

\midrule
\multicolumn{11}{c}{\textit{Average Token Reduction} $\downarrow 66.7\%$} \\
\midrule
\rowcolor{baserow}
FastV
& ECCV'24
& 49.8 & \second{55.1} & \second{54.6}
& 42.3 & \second{47.6} & \second{46.6}
& \second{53.5} & \second{53.6}
& \second{59.1\%} \\

\rowcolor{baserow}
PDrop
& CVPR'25
& \best{63.4} & \best{64.8} & \best{60.6}
& \best{55.5} & \best{57.3} & \best{51.2}
& \best{61.6} & \best{61.0}
& \best{69.5\%} \\

\midrule
\rowcolor{routerow}
FastV+Reroute
& Ours
& 46.8 & 50.8 & 43.0
& 39.2 & 43.8 & 35.5
& 44.0 & 43.5
& 50.6\% \\

\rowcolor{routerow}
PDrop+Reroute
& Ours
& \second{58.1} & 53.8 & 46.8
& \second{50.5} & 47.2 & 39.0
& 48.7 & 48.0
& 57.2\% \\

\midrule
\multicolumn{11}{c}{\textit{Average Token Reduction} $\downarrow 77.8\%$} \\
\midrule
\rowcolor{baserow}
FastV
& ECCV'24
& 27.4 & 33.9 & \second{31.5}
& 22.6 & 29.3 & \best{28.1}
& \second{31.5} & \best{31.9}
& 34.7\% \\

\rowcolor{baserow}
PDrop
& CVPR'25
& \best{37.1} & \best{40.5} & \best{32.0}
& \best{31.6} & \best{34.1} & \second{27.4}
& \best{32.1} & \second{31.5}
& \best{38.9\%} \\

\midrule
\rowcolor{routerow}
FastV+Reroute
& Ours
& 27.0 & 30.4 & 24.9
& 22.1 & 24.8 & 20.2
& 24.0 & 23.6
& 28.8\% \\

\rowcolor{routerow}
PDrop+Reroute
& Ours
& \second{36.3} & \second{36.2} & 29.6
& \second{30.3} & \second{30.7} & 24.2
& 29.8 & 29.2
& \second{36.1\%} \\

\midrule
\multicolumn{11}{c}{\textit{Average Token Reduction} $\downarrow 88.9\%$} \\
\midrule
\rowcolor{baserow}
FastV
& ECCV'24
& 6.3 & 7.6 & 4.2
& 4.1 & 4.1 & 3.5
& 5.1 & 5.3
& 5.9\% \\

\rowcolor{baserow}
PDrop
& CVPR'25
& \second{11.9} & \second{15.0} & \second{8.9}
& \second{10.6} & \second{12.3} & \second{8.1}
& \second{10.7} & \second{10.5}
& \second{12.9\%} \\

\midrule
\rowcolor{routerow}
FastV+Reroute
& Ours
& 6.6 & 8.2 & 4.5
& 3.9 & 4.4 & 3.9
& 5.2 & 5.5
& 6.1\% \\

\rowcolor{routerow}
PDrop+Reroute
& Ours
& \best{17.8} & \best{21.5} & \best{15.8}
& \best{14.1} & \best{17.7} & \best{12.9}
& \best{15.9} & \best{15.1}
& \best{19.1\%} \\

\bottomrule
\end{tabular}%
}

\vspace{0.35em}
\footnotesize{
\colorbox{vanillarow}{Vanilla baseline};
\colorbox{baserow}{Base methods};
\colorbox{routerow}{Reroute variants}. \\
For RefCOCO and RefCOCO+, A/B denote testA/testB.
Avg. Ratio is the mean score ratio against the vanilla baseline,
computed over subsets where both scores are available.
Schedule details are reported in the appendix.
}
\end{table}

In the moderate-reduction regimes, the early PDrop schedule remains the strongest grounding configuration: at $66.7\%$ token reduction, PDrop (L2) reaches a $69.5\%$ average ratio, while PDrop + Reroute reaches $57.2\%$; at $77.8\%$ reduction, PDrop reaches $38.9\%$, while PDrop + Reroute reaches $36.1\%$. Under the most aggressive $88.9\%$ reduction, however, PDrop + Reroute improves over PDrop from $12.9\%$ to $19.1\%$, suggesting that recoverable deferral becomes useful when irreversible pruning is most likely to remove essential grounding evidence.

We therefore use Qwen3-VL as a case study showing that Reroute can help under extreme compression, but that training-free schedules should still be chosen with model-specific architectural constraints in mind, especially for backbones with additional visual-feature injection mechanisms.

\subsection{Additional qualitative results}
\label{app:qualitative-results}

\begin{figure}[t]
    \centering
    \includegraphics[width=\linewidth]{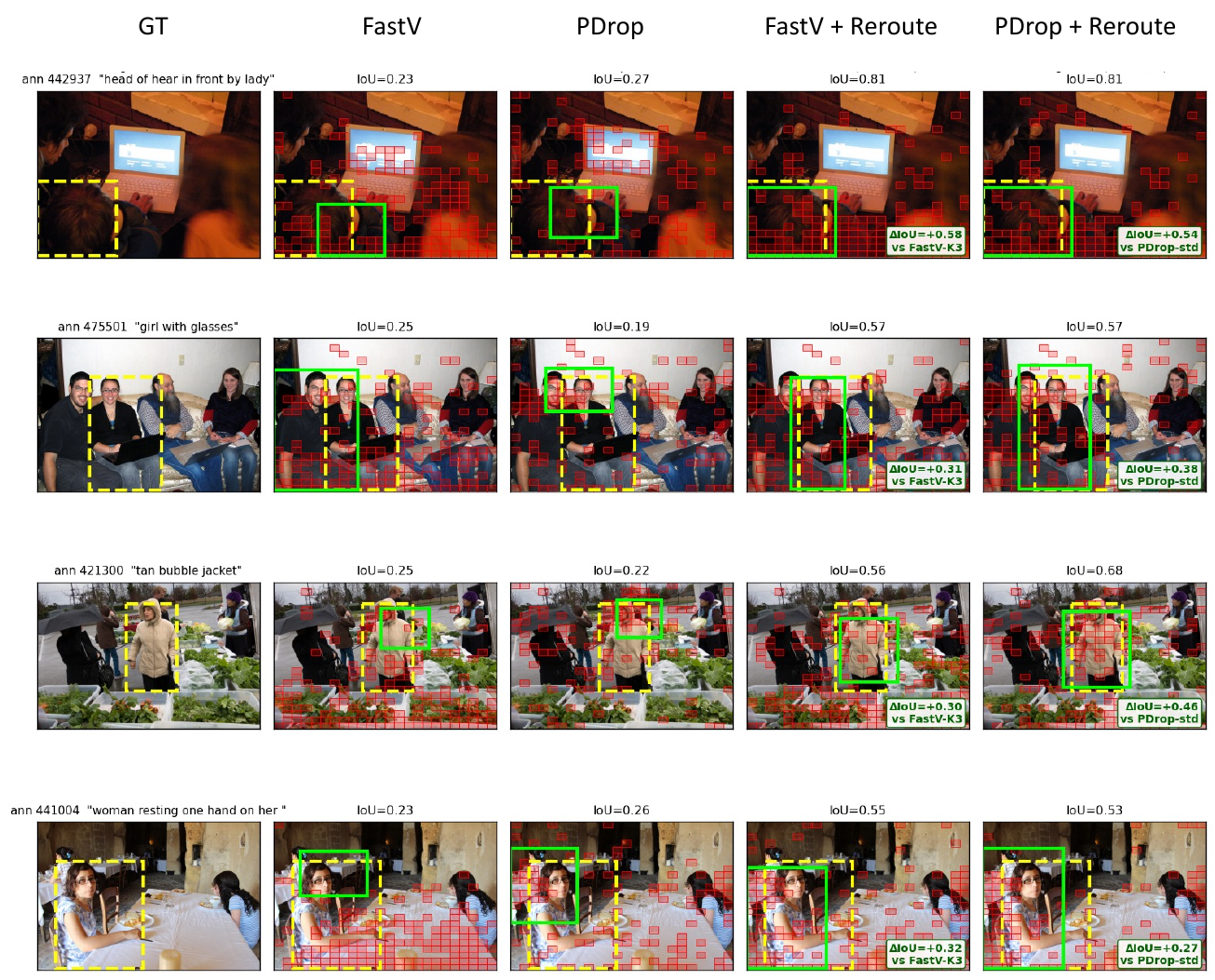}
    \caption{Selected visual-token masks on LLaVA-1.5. The highlighted regions indicate the visual tokens selected by the routing policy under the corresponding token budget.}
    \label{fig:supp_llava_mask}
\end{figure}

\begin{figure}[t]
    \centering
    \includegraphics[width=\linewidth]{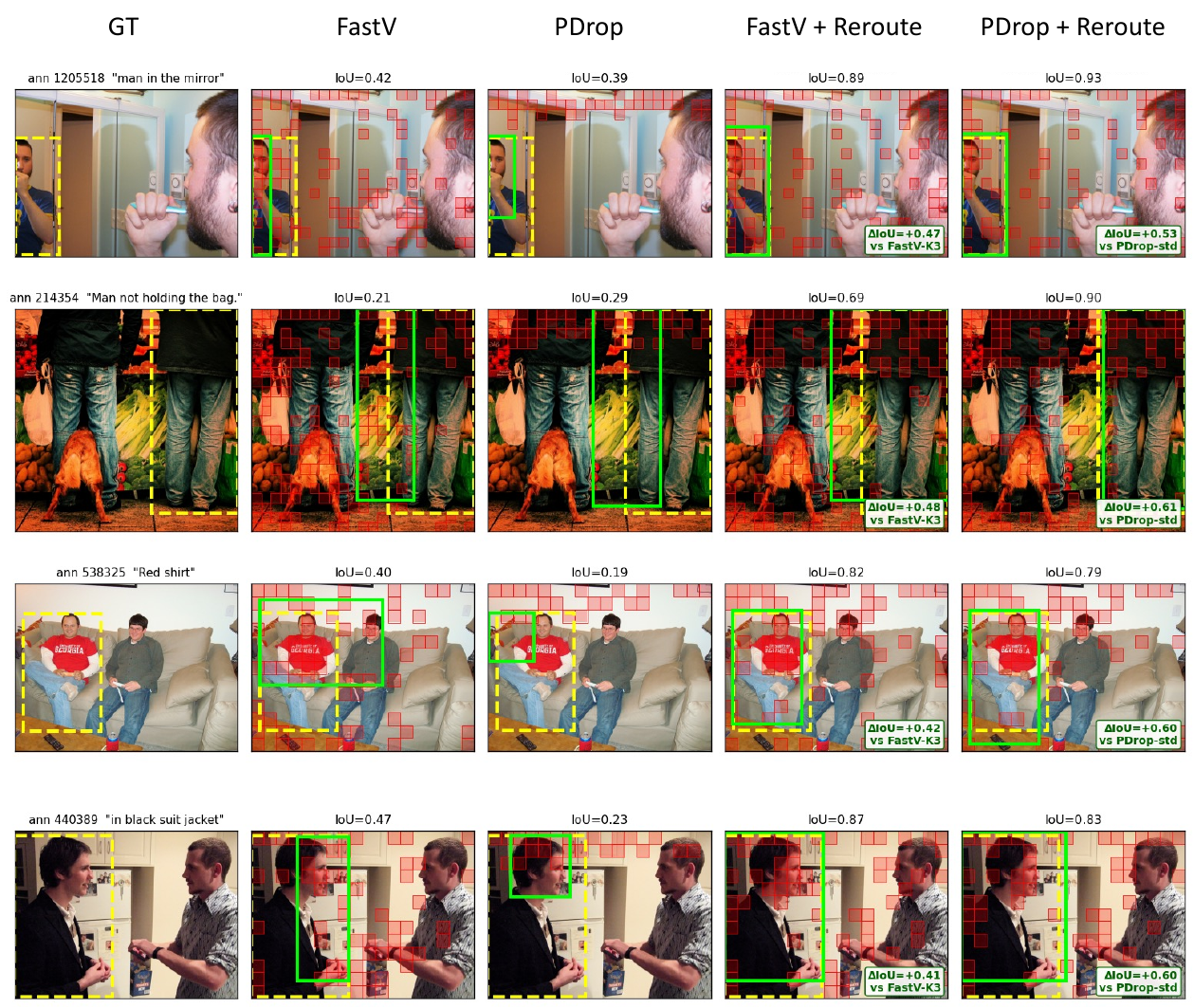}
    \caption{Selected visual-token masks on Qwen2.5-VL. The highlighted regions indicate the visual tokens selected by the routing policy under the corresponding token budget.}
    \label{fig:supp_qwen25_mask}
\end{figure}

\begin{figure}[p]
    \centering
    \includegraphics[
        height=0.78\textheight,
        keepaspectratio
    ]{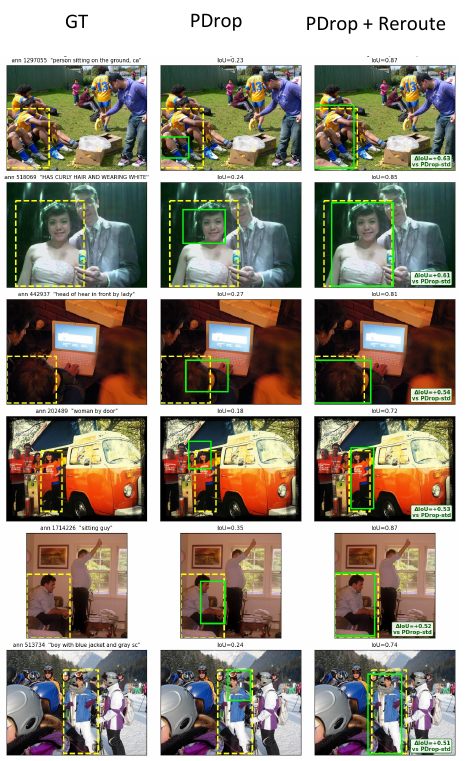}
    \caption{
    Additional RefCOCO-family grounding examples on LLaVA-1.5. We compare
    pruning baselines with their corresponding Reroute variants under matched
    token budgets.
    }
    \label{fig:supp_llava15_bbox}
\end{figure}

\begin{figure}[t]
    \centering
    \includegraphics[width=\linewidth]{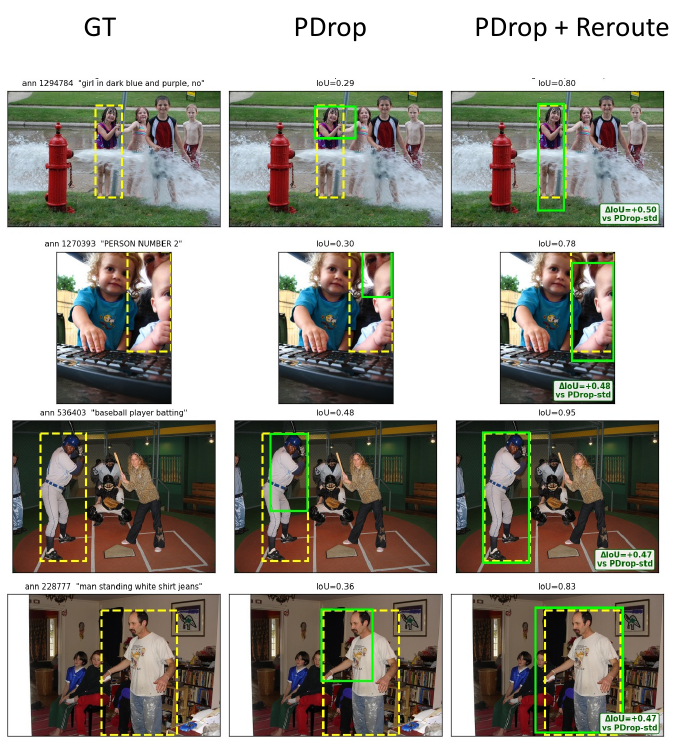}
    \caption{Additional RefCOCO-family grounding examples on Qwen2.5-VL. Reroute preserves deferred visual evidence and improves grounding over matched pruning baselines.}
    \label{fig:supp_qwen25_bbox}
\end{figure}

\begin{figure}[t]
    \centering
    \includegraphics[width=\linewidth]{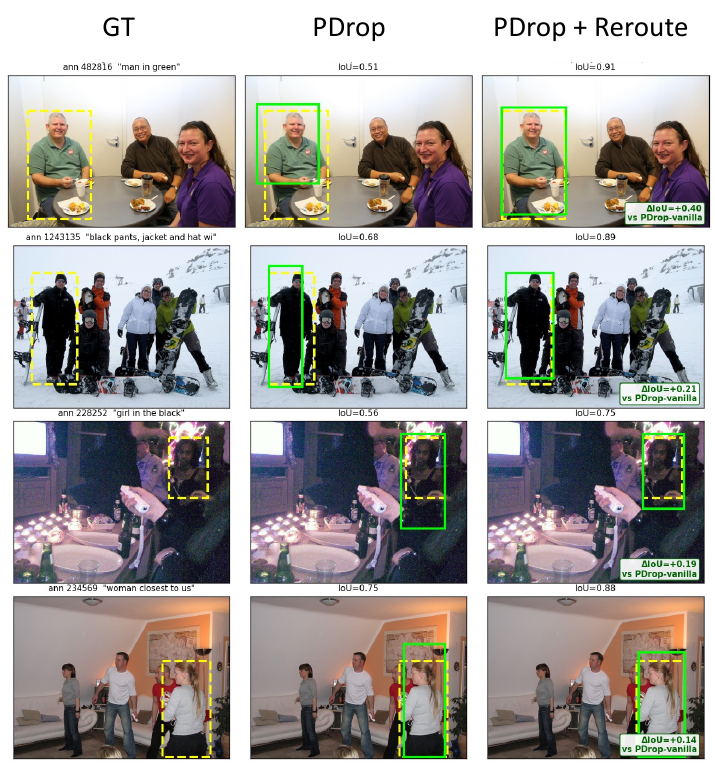}
    \caption{Additional RefCOCO-family grounding examples on Qwen3.5. We compare pruning baselines and their Reroute variants under matched gated-attention token budgets.}
    \label{fig:supp_qwen35_bbox}
\end{figure}

We provide additional qualitative visualizations for RefCOCO-family grounding in Figures~\ref{fig:supp_llava15_bbox}, \ref{fig:supp_qwen25_bbox}, and \ref{fig:supp_qwen35_bbox}. These examples cover LLaVA-1.5, Qwen2.5-VL, and Qwen3.5, and compare pruning baselines with their corresponding Reroute variants under matched token budgets. We further visualize selected visual-token masks for LLaVA-1.5 and Qwen2.5-VL in Figures~\ref{fig:supp_llava_mask} and \ref{fig:supp_qwen25_mask}, showing which image regions remain active under the routing policy.

\end{document}